\documentclass[10pt,shortpaper,twoside,web,logowidth=10.5]{ieeecolor}
\usepackage{generic}
\usepackage{cite}
\usepackage{amsmath,amssymb,amsfonts}
\usepackage{graphicx}
\usepackage{textcomp}
\usepackage{stfloats}
\usepackage{graphicx}
\usepackage{booktabs}
\usepackage{float}
\usepackage{amssymb,amsfonts}
\usepackage{listings}
\usepackage{url}
\usepackage[ruled,vlined]{algorithm2e}
\usepackage{algpseudocode}
\usepackage{mathrsfs}
\usepackage{amsmath,bm}
\usepackage[switch]{lineno}
\usepackage[table,xcdraw]{xcolor}
\usepackage{scalerel}
\usepackage{tikz}
\usepackage{multirow}
\usepackage{tipa}

\usepackage{url}
\usepackage{balance}
\usepackage{graphicx}
\usepackage{amsfonts}
\usepackage{fancyhdr}
\usepackage{comment}
\usepackage{times}
\usepackage{amsmath}
\usepackage{changepage}
\usepackage{amssymb}
\usepackage{enumerate}
\usepackage{bm}
\usepackage{subfigure}
\usepackage{calligra}
\usepackage{multirow}
\usepackage{tabularx}
\usepackage{booktabs}
\usepackage{mathtools}
\usepackage{arydshln}
\usepackage{latexsym}
\usepackage{amsmath}
\usepackage{amssymb}
\usepackage{color}
\usepackage{soul}
\usepackage{gensymb}
\usepackage{cite}
\usepackage{color}
\usepackage{cuted}
\usepackage{etoolbox}
\usepackage{capt-of}
\usepackage[flushleft]{threeparttable}
\usepackage{ulem}
\usepackage{float}
\usepackage{booktabs}
\usepackage{bookmark}
\usepackage{hyperref}
\usetikzlibrary{svg.path}

\definecolor{orcidlogocol}{HTML}{A6CE39}
\tikzset{
  orcidlogo/.pic={
    \fill[orcidlogocol] svg{M256,128c0,70.7-57.3,128-128,128C57.3,256,0,198.7,0,128C0,57.3,57.3,0,128,0C198.7,0,256,57.3,256,128z};
    \fill[white] svg{M86.3,186.2H70.9V79.1h15.4v48.4V186.2z}
                 svg{M108.9,79.1h41.6c39.6,0,57,28.3,57,53.6c0,27.5-21.5,53.6-56.8,53.6h-41.8V79.1z M124.3,172.4h24.5c34.9,0,42.9-26.5,42.9-39.7c0-21.5-13.7-39.7-43.7-39.7h-23.7V172.4z}
                 svg{M88.7,56.8c0,5.5-4.5,10.1-10.1,10.1c-5.6,0-10.1-4.6-10.1-10.1c0-5.6,4.5-10.1,10.1-10.1C84.2,46.7,88.7,51.3,88.7,56.8z};
  }
}
\newcommand\orcidicon[1]{\href{https://orcid.org/#1}{\mbox{\scalerel*{
\begin{tikzpicture}[yscale=-1,transform shape]
\pic{orcidlogo};
\end{tikzpicture}
}{|}}}}

\usepackage{hyperref}
\hypersetup{
    colorlinks=true,
    linkcolor=black,
    filecolor=black,      
    urlcolor=black,
    citecolor=black
}

\def\BibTeX{{\rm B\kern-.05em{\sc i\kern-.025em b}\kern-.08em
    T\kern-.1667em\lower.7ex\hbox{E}\kern-.125emX}}
\newcommand{\dgt}{\textcolor[rgb]{0.4,0.4,0.4}}

\begin{document}
\title{\LARGE{LCE-Calib: Automatic LiDAR-Frame/Event Camera Extrinsic Calibration With A Globally Optimal Solution}}

\author{\large{Jianhao Jiao$^{2,4}$, \textit{IEEE Member}, Feiyi Chen$^{2}$, Hexiang Wei$^{2}$, Jin Wu$^{2}$, \textit{IEEE Member}, \\Ming Liu$^{1,2,3}$, \textit{IEEE Senior Member}}
  \thanks{This work was supported by Guangdong Basic and Applied Basic Research Foundation, under project 2021B1515120032, Foshan-HKUST Project no. FSUST20-SHCIRI06C, and Project of Hetao Shenzhen-Hong Kong Science and Technology Innovation Cooperation Zone (HZQB-KCZYB-2020083), awarded to Prof. Ming Liu.}
  \thanks{$^{1}$Robotics and Autonomous Systsms, The Hong Kong University of Science and Technology (Guangzhou), Nansha, Guangzhou, 511400, Guangdong, China.}
  \thanks{$^{2}$Department of Electronic and Computer Engineering, Hong Kong University of Science and Technology, Hong Kong, China. \{jjiao,hweiak,fchenak,jwucp\}@connect.ust.hk, \{eelium\}@ust.hk.}
  \thanks{$^{3}$HKUST Shenzhen-Hong Kong Collaborative Innovation Research Institute, Futian, Shenzhen.}
  \thanks{$^{4}$CWB Institute of Autonomous Driving, Nanshan, Shenzhen.}
  \thanks{\textit{Corresponding author: Ming Liu.}}
}

\maketitle

\begin{abstract}
  The combination of LiDARs and cameras enables a mobile robot to perceive environments with multi-modal data, becoming a key factor in achieving robust perception. Traditional frame cameras are sensitive to changing illumination conditions, motivating us to introduce novel event cameras to make LiDAR-camera fusion more complete and robust. However, to jointly exploit these sensors, the challenging extrinsic calibration problem should be addressed.

  This paper proposes an automatic checkerboard-based approach to calibrate extrinsics between a LiDAR and a frame/event camera, where four contributions are presented. Firstly, we present an automatic feature extraction and checkerboard tracking method from LiDAR’s point clouds. Secondly, we reconstruct realistic frame images from event streams, applying traditional corner detectors to event cameras. Thirdly, we propose an initialization-refinement procedure to estimate extrinsics using point-to-plane and point-to-line constraints in a coarse-to-fine manner. Fourthly, we introduce a unified and globally optimal solution to address two optimization problems in calibration. Our approach has been validated with extensive experiments on 19 simulated and real-world datasets and outperforms the state-of-the-art.
\end{abstract}


\section{Introduction}
\label{sec:introduction}

\subsection{Motivation}
One of the challenges to realizing autonomous mobile robots should be the perception problem.
As the front-end module of a robotic system, perception offers essential information to high-level navigation.
Nowadays, the fusion of a LiDAR and a camera, called \textit{LC-Fusion}, provides a promising solution for robust perception.
LC-Fusion is able to overcome individual limitations of both LiDARs and cameras, producing confident results to boost various tasks such as simultaneous localization and mapping (SLAM) \cite{lin2022r,yu2022accurate} and pattern recognition \cite{qi2018frustum}.
Overall, LC-Fusion enjoys the following advantages:
\begin{itemize}
    \item \textbf{Accessibility}:
          Benefiting from the rapid development of sensory technologies, both cameras and LiDARs are portable and accessible for various commercial mobile robots, including drones \cite{zhang2019maximum}, quadrupedal robots \cite{jiao2022fusionportable}, and self-driving cars \cite{cao2021learning}.
    \item \textbf{Information Sufficiency}:
          LC-Fusion provides multi-modal information that is typically sufficient for robot navigation.
          Specifically, LiDARs directly offer sparse but accurate 3D measurements of surrounding objects.
          In contrast, cameras capture dense and high-resolution 2D images, which is beneficial to object recognition.
    \item \textbf{Wide Usage}:
          Research on LiDARs, cameras, and their fusion has attracted much attention from the research community.
          Algorithms targeting at different perception tasks have immediate solutions and are applicable to many robotic navigation tasks.
\end{itemize}

\begin{figure}[t]
    \centering
    \subfigure[]
    {\label{fig:lidar_garden}\centering\includegraphics[width=0.110\textwidth]{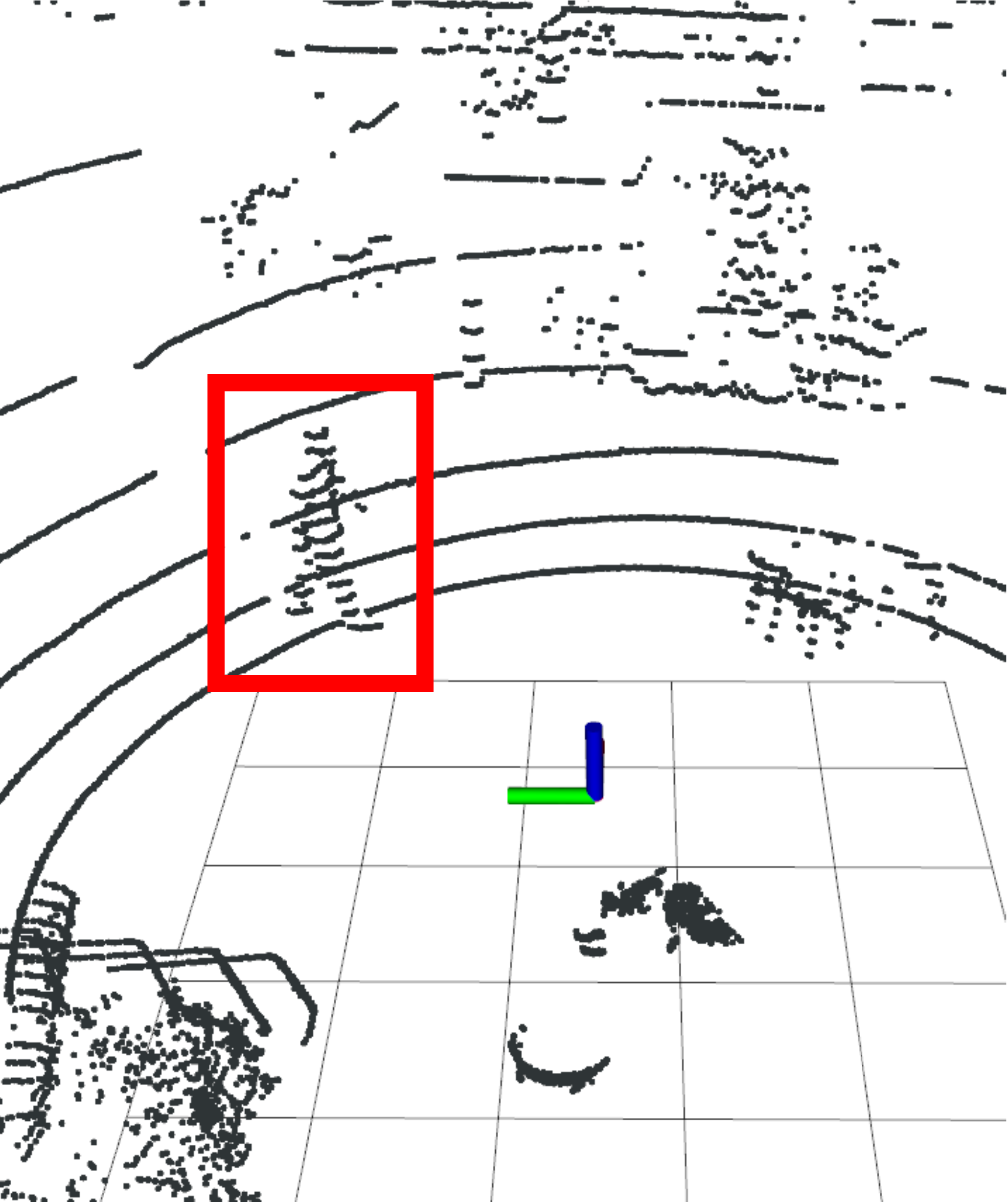}}
    \subfigure[]
    {\label{fig:image_garden}\centering\includegraphics[width=0.115\textwidth]{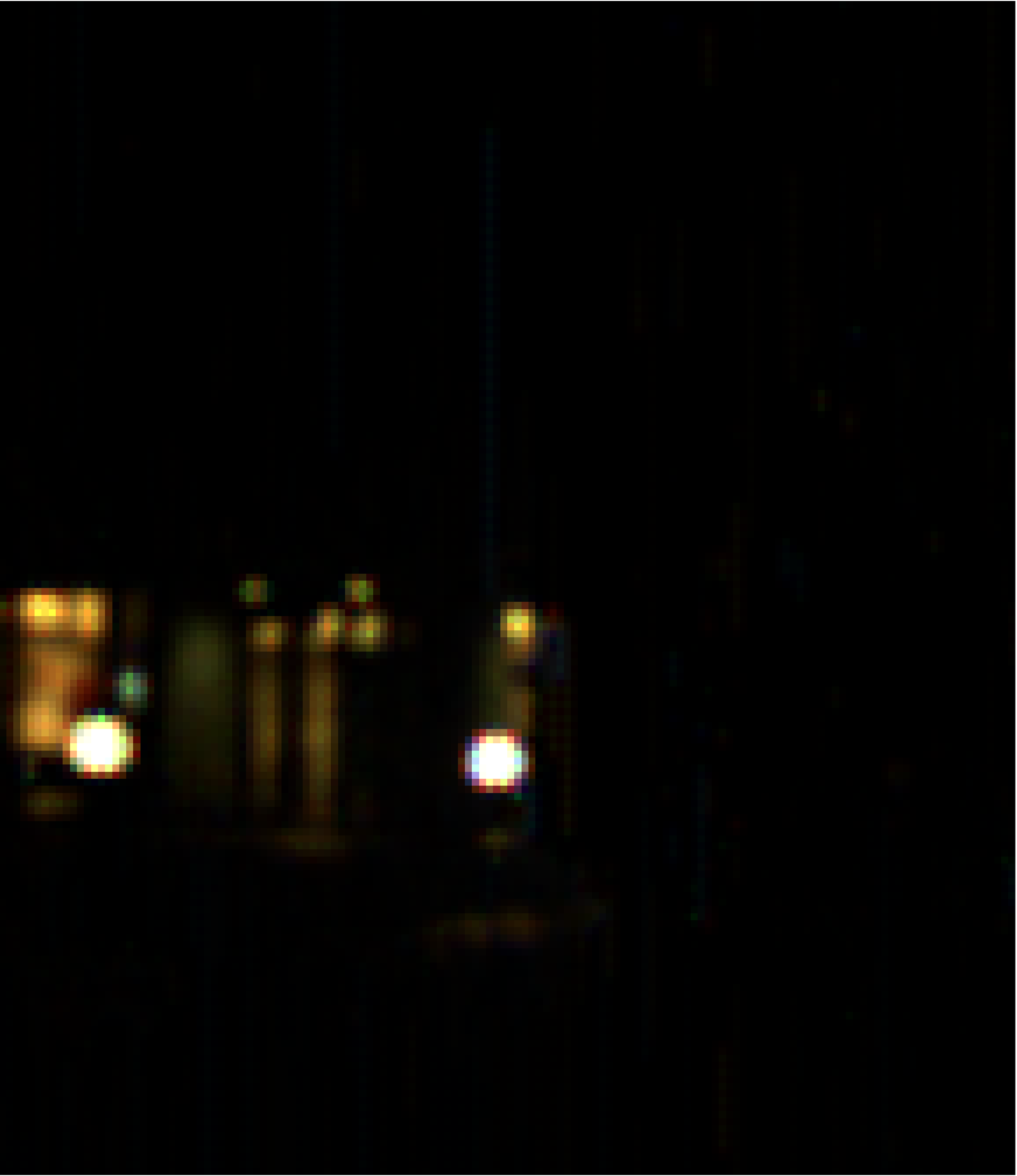}}
    \subfigure[]
    {\label{fig:image_event_garden}\centering\includegraphics[width=0.115\textwidth]{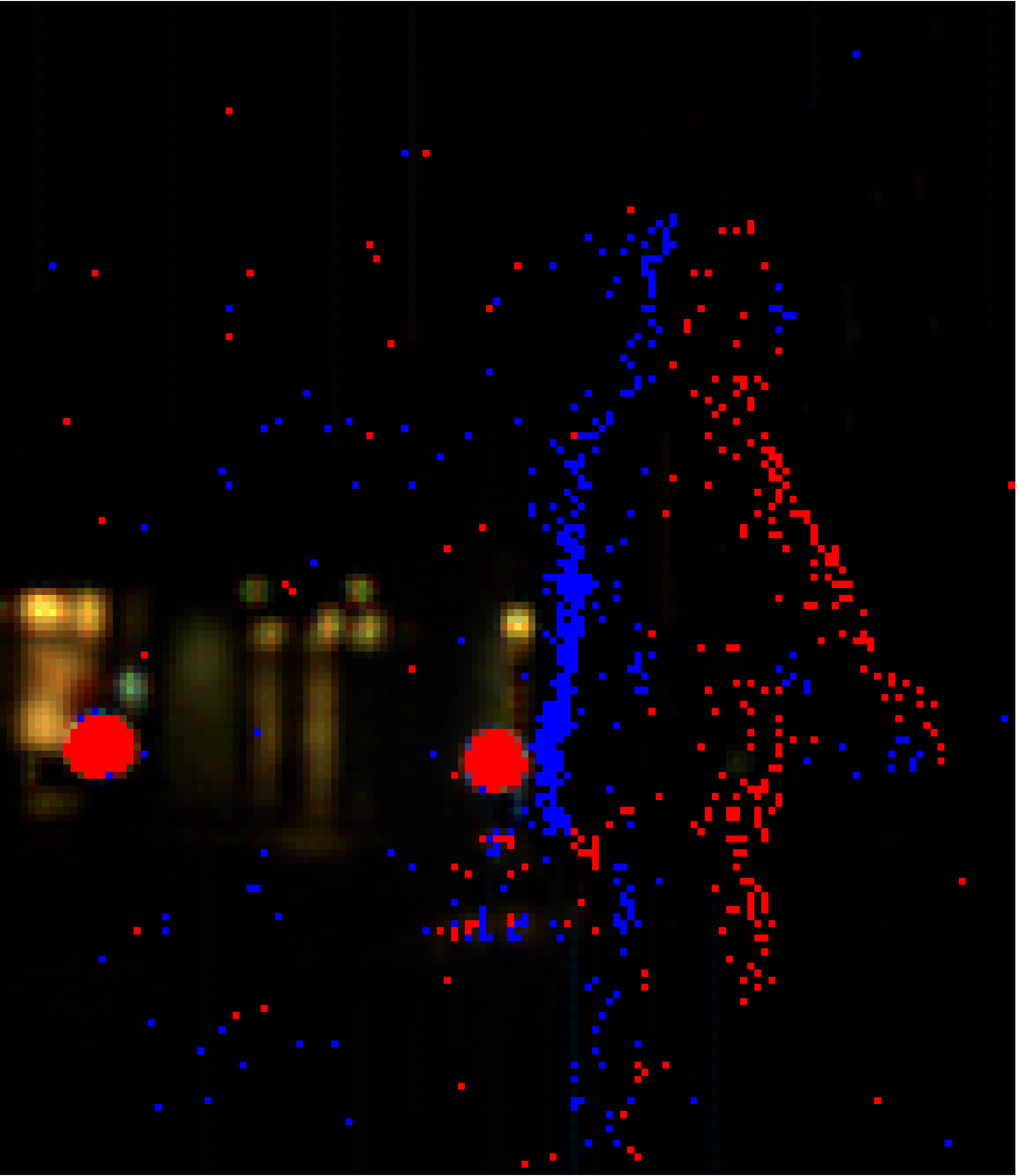}}
    \subfigure[]
    {\label{fig:image_event_garden}\centering\includegraphics[width=0.115\textwidth]{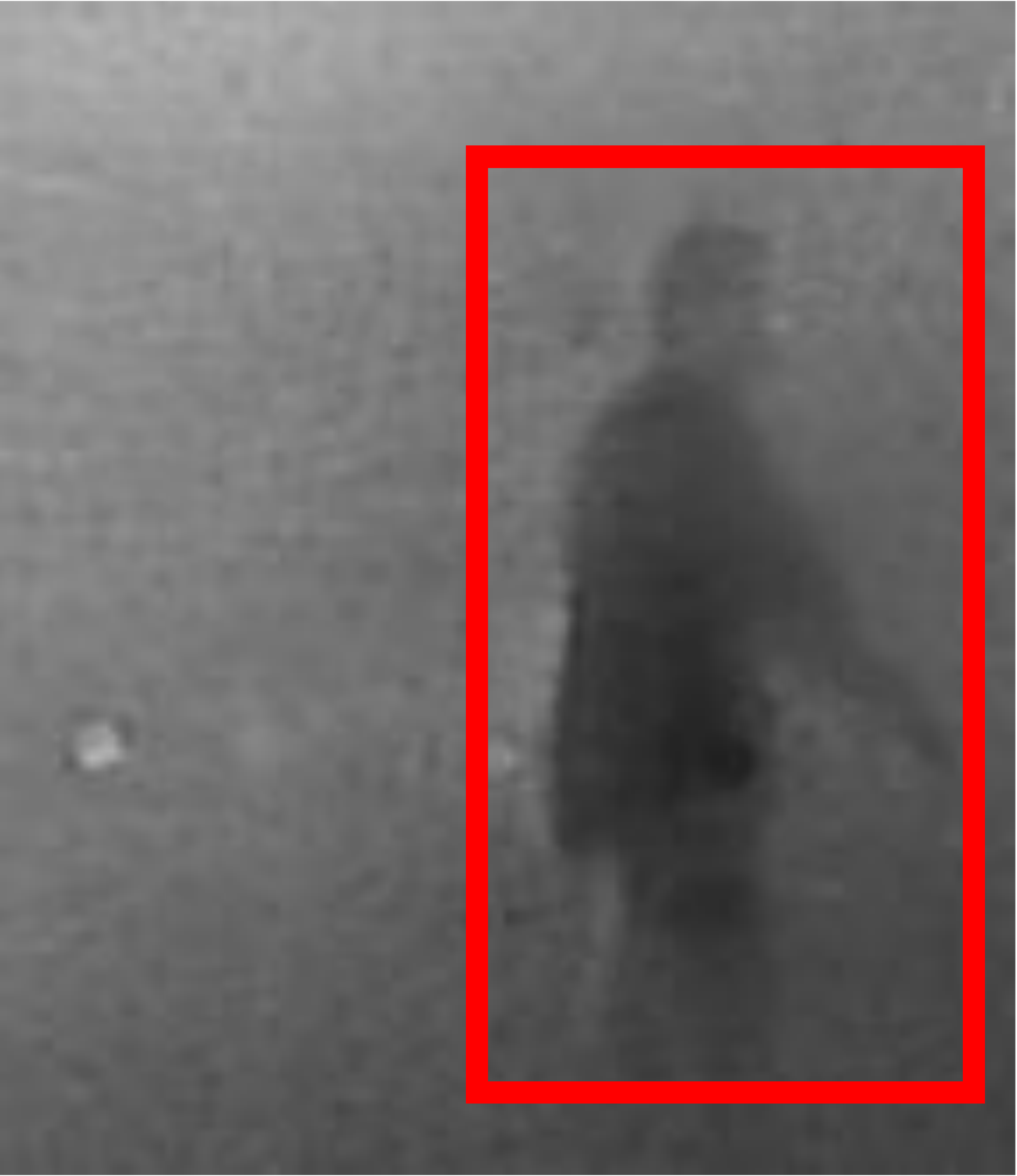}}
    \caption{Sensor measurements in a garden at night:
        (a) a LiDAR point cloud,
        (b) a frame image,
        (c) events that are printed on a frame image, and
        (d) reconstructed images using events.
        We can distinguish a human from the point cloud and event image except for the frame image. This indicates that the frame image is sensitive to the weak light. Please refer to the color version for the better visualization.}
    \label{fig:sensor}
\end{figure}

However, traditional frame cameras are commonly sensitive to changing illumination (e.g., darkness and glare).
Cameras cannot capture scene information completely and may fail several vision-based algorithms.
LC-Fusion may degenerate into a LiDAR-only configuration since most images are noisy.
This issue motivates us to explore a novel type of sensor, event cameras \cite{gallego2020event}, to complement the traditional LiDAR-camera fusion in challenging environments.
The event camera augments the original sensor setup, and we can fuse three types of sensor input for perception.
This new sensor fusion mode is called \textit{LCE-Fusion}.

Event cameras are bio-inspired sensors.
Different from frame cameras that capture images at a fixed rate, event cameras asynchronously capture the per-pixel \textit{intensity changes} and output a stream of \textit{2D events}.
Each event is encoded with information, including the triggered time, pixel localization, and the sign of the intensity change.
Event cameras have high temporal resolution ($\mu s$-level), high dynamic range ($140$dB v.s. $60$dB of frame cameras), and low power consumption.
They have great potential for several computer vision and robotic tasks, e.g.,
high-speed motion estimation \cite{bryner2019event} and high dynamic range perception \cite{rebecq2019high}.
Fig. \ref{fig:sensor} visualizes the enhancement brought by an event camera in a dark garden.


\subsection{Challenges}
The extrinsic calibration, estimating the relative rotational and translational offset from the reference frame to the target frame, is an indispensable step in using the LCE-Fusion.
Our goal is to design a general and automatic extrinsic calibration approach for the LCE-Fusion. However, as emphasized in \cite{choi2015extrinsic}, challenges of automatic calibration arise from the three aspects: feature extraction given noisy data, data association across multi-modal sensors, and parameter estimation.

A standard checkerboard is desirable in calibration, offering distinctive features (e.g., corners, boundaries, and a plane) and known geometry for feature matching. Although existing methods \cite{verma2019automatic} have demonstrated the validity of this paradigm, several issues, including feature extraction, data association, event representation, and global optimization, have not been addressed well:
\begin{itemize}
    \item \textbf{Automatic Feature Extraction}: It is straightforward to detect the checkerboard from images by the off-the-shelf softwares such as OpenCV. However, this is not applicable to point clouds whose data model is fundamentally different. The detection of the checkerboard from point clouds is a nontrivial problem and is particularly hard if point clouds contain doors, tables, and walls with the planar shape. Several works \cite{verma2019automatic} have to finish the feature extraction manually.
    \item \textbf{Automatic Feature Matching}: An automatic approach to match checkerboard features between LiDARs and cameras are needed. However, this raises the issue that the symmetric shape of the checkerboard may lead to ambiguous data association, resulting in suboptimal or unreliable extrinsics.
    \item \textbf{Event Representation}: The asynchronous and sparse nature of event cameras makes the feature selection difficult. We need a method to convert a group of events into an image-type representation since traditional methods can be directly applied.
    \item \textbf{Globally Optimal Solution}: The extrinsic optimization problem is generally non-convex. This implies that the typical Gauss-Newton solution that linearizes the objective is approximate to the globally optimal solution. Thus, introducing a globally optimal solver in calibration given noisy sensor measurements is important.
\end{itemize}

\subsection{Contributions}
To tackle these challenges, we propose \textit{LCE-Calib}, a unified extrinsic calibration method for the LCE sensor configuration.
In overall, LCE-Calib presents the following \textit{contributions}.
\begin{enumerate}
    \item We propose an automatic checkerboard extraction and tracking method that is robust to external noisy objects from point clouds (see Section \ref{sec:methodology_fsl}). This method is also noise-aware, since we reduce the bias of LiDAR points by projecting points onto a reference plane and model the uncertainty of normal vectors.
    \item We introduce the learning-based approach to reconstruct frame images from event streams for the downstream extrinsic calibration task. The resulting images allow the usage of traditional corner detectors (see Section \ref{sec:methodology_ire}).
    \item We design the calibration process with an initialization-refinement philosophy to utilize point-to-plane and point-to-line constraints in a coarse-to-fine manner. This avoids the ambiguity issue in data association caused by the board's symmetric shape (see Section \ref{sec:ext_calibration}).
    \item We introduce a general solver to globally solve two optimization problems in calibration: the Perspective-n-Point (PnP) and Point-to-Plane registration (PtPL) problems. The benefits from it is that the resulting extrinsics are accurate even if measurements are noisy.
\end{enumerate}

The proposed method is evaluated extensively on different sensor devices in various calibration scenes from indoor offices to outdoor grounds. The proposed LCE-Calib outperforms the state-of-the-art (SOTA) calibration method, achieving an extrinsic accuracy of centimeters in translation and deci-degrees in rotation. To benefit the community, we publicly release the experimental data and code.\footnote{\url{https://github.com/HKUSTGZ-IADC/LCECalib}}

\section{Related Work}
\label{sec:related_work}

\begin{figure*}[t]
  \centering
  \includegraphics[width=0.85\textwidth]{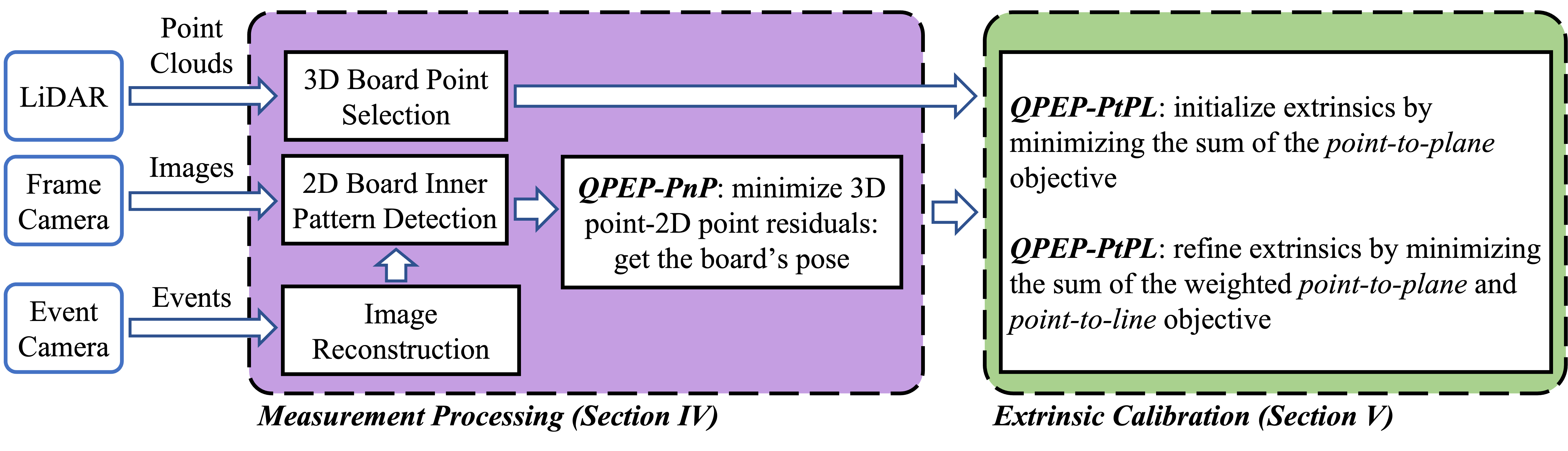}
  \caption{Block diagram illustrating the full pipeline of the proposed LCE-Calib method.
    The method first processes raw sensor measurements (see Section \ref{sec:measure_process}) to extract features and then initializes as well as refines sensors' extrinsics (see Section \ref{sec:ext_calibration}).}
  \label{fig:methodology_overview}
\end{figure*}

Sensor calibration is a fundamental problem in robotics.
This section briefly reviews calibration results on the LiDAR-camera sensor suite and event cameras.

\subsection{LiDAR-Camera Calibration}
Related approaches are mainly divided into marker-based and marker-less approaches, depending on whether artificial targets are used or not during the process.

\subsubsection{Marker-based Methods}
Although marker-based calibration requires artificial targets to be observable in the scene and sometimes needs manual intervention, such methods are still prevalent due to the higher accuracy and robustness compared with nontarget-based calibration \cite{huang2018geometric}.
They benefit from the known geometry information of the target, imposing sufficient and reliable geometric constraints to estimate extrinsics.

Geometric solids \cite{guennebaud2010eigen}, checkerboards \cite{wu2021simultaneous,zhou2018automatic}, and polygons \cite{liao2018extrinsic} are the widely used targets.
But among them, checkerboards are the most common targets since they also help the cameras' intrinsic calibration \cite{zhang2000flexible}.
Zhang and Pless \cite{zhang2004extrinsic} first introduced the checkerboard into the extrinsic calibration of a perspective camera and a 2D laser rangefinder. They placed the checkerboard at multiple poses to gather sufficient constraints induced by plane-line correspondences.
Unnikrishnan \textit{et al.} \cite{unnikrishnan2005fast} extended this method to the extrinsic calibration with a 3D LiDAR.
They estimated the planar parameters of the board from the LiDAR and camera. They used plane-plane correspondences to initialize rotation and translation decoupled while minimizing the point-plane distance to refine the LiDAR-camera transformation.
Our method also employs this way to initialize extrinsics.
Pandey \textit{et al.} \cite{pandey2010extrinsic} addressed the omnidirectional camera-LiDAR calibration problem,
Mirzaei \textit{et al.} \cite{mirzaei20123d} additionally investigated the LiDAR intrinsic model,
and Koo \textit{et al.} \cite{koo2020analytic} analytically derived the uncertainty of plane parameters and explored its effect in calibration.

Besides plane information, the boundaries of objects and corner features also offer strong constraints.
Sim \textit{et al.} \cite{sim2016indirect} determined lines from two non-coplanar surfaces of a V-shaped target and exploited linear constraints to estimate extrinsics.
Moghadam \textit{et al.} \cite{moghadam2013line} extracted natural linear features in the scene,
while Zhou \textit{et al.} \cite{zhou2018automatic} jointly exploited 3D line-to-line and point-to-plane correspondences to establish constraints.
Finally, Huang \textit{et al.} \cite{huang2020improvements} extracted corner features that are the intersections of two boundaries from an Apriltag \cite{olson2011apriltag} and then solved the PnP problem \cite{lepetit2009epnp}.

Our checkerboard-based approach presents several new features from following aspects:
\textit{1)} features from images and point clouds are automatically extracted;
\textit{2)} board points are continuously tracked to improve the detection success rate and reduce processing time at the next frame;
\textit{3)} both planar and line features are fully utilized in different stages of calibration;
\textit{4)} an event camera is also extrinsically calibrated with the LiDAR;
\textit{5)} the uncertainty information is represented with the Lie group-based formulation \cite{barfoot2014associating}; and
\textit{6)} a new globally optimal solver is introduced.

\subsubsection{Marker-less Methods}
Marker-less calibration methods search for correspondences between geometric features found in natural environments, such as lines, edges, and planar regions. They do not rely on explicit shapes from known targets, having great flexibility for field robots.

Levinson \textit{et al.} \cite{levinson2013automatic} put forward the first online calibration approach for a LiDAR-camera system by optimally aligning 3D edge points with image contours.
Different metrics based on the planar information \cite{chen2022pbacalib}, edges \cite{yuan2021pixel}, and semantic constraints \cite{yoon2021targetless} were also proposed.
Meanwhile, Shi \textit{et al.} \cite{shi2020calibrcnn} proposed an end-to-end neural network to calibrate the extrinsics.
However, the accuracy of markerless methods is commonly inferior to marker-based methods if calibration scenes are non-ideal.
In our approach, we show that the checkerboard benefits the intrinsic calibration of cameras. We also investigate the checkerboard-based calibration with an extension to event cameras.

\subsection{Event Camera Calibration}
The unique characteristics of event cameras raise the demand for novel solutions to address the calibration problem.
Preliminary works on event cameras have designed blinking LED patterns \cite{dominguez2019bio}, or screens \cite{mueggler2014event} as the calibration target.
The rapid illumination change triggers events and the patterns can be detected even if the cameras are static.
Once features are extracted, traditional optimization-based calibration backends can be used.
Nevertheless, these approaches may have large motion blur and noisy events \cite{muglikar2021calibrate}.
Recent works are able to replace the custom-built calibration checkerboards with the standard ones by implementing one of two techniques:
\textit{1)} several event cameras output synchronous frame images \cite{gallego2020event};
\textit{2)} high-quality frame images are reconstructed from event streams by utilizing the sensing characteristics of event cameras \cite{muglikar2021calibrate}.
The latter approaches should be more applicable to more types of event cameras that only have an event output. They also benefit from recent advancements in image reconstruction from events \cite{rebecq2019high}.

Our method follows the second technique in event-based calibration.
We further propose a unified method to calibrate extrinsics between a LiDAR and an event camera based on these works. The method is shown with high accuracy under detailed evaluation.

\section{Problem Statement}
\label{sec:problem_statement}

The extrinsic calibration problem is divided into three subproblems:
PnP, PtPL, and Point-to-Line (or Edge) (PtL) registration problems.
All of them are generally formulated as quadratic pose estimation problems (QPEPs) \cite{wu2022quadratic}.
Before delving into the details of LCE-Calib, we first introduce basic concepts here.
Section \ref{sec:ps_notation} presents notations.
Section \ref{sec:ps_qpep} introduces the definition of QPEP and general idea to solve QPEPs based on our previous work.

\subsection{Notations and Definitions}
\label{sec:ps_notation}
We consider a sensor suite that consists of a LiDAR, a frame camera, and an event camera.
Frames of these sensors are defined as $()^{l}$, $()^{c}$, and $()^{e}$ respectively.
We also define $()^{b}$ as the frame of the checkerboard, where the origin stays at the board's center, and the $z$-axis is perpendicular to the board's plane.
We use $\bm{t}\in\mathbb{R}^{3}$ and $\bm{R}\in SO(3)$ to represent the 3D translation and rotation. Especially, the rotation matrix is from the Lie group $SO(3)$ where $\bm{R}^{\top}\bm{R}=\bm{I}$, $\det\bm{R}=1$.
For any real 3D vector $\bm{\phi}\in \mathbb{R}^{3}$, its skew-symmetric matrix is
\begin{equation}
  \bm{\phi}^{\wedge}
  =
  \begin{bmatrix}
    0         & -\phi_{3} & \phi_{2}  \\
    \phi_{3}  & 0         & -\phi_{1} \\
    -\phi_{2} & \phi_{1}  & 0         \\
  \end{bmatrix}
  \in
  \mathfrak{so}(3),
\end{equation}
which is an element from the Lie algebra $\mathfrak{so}(3)$.
We use the exponential operator to associate an element from $SO(3)$ with an element from $\mathfrak{so}(3)$:
$\bm{R}=\exp(\bm{\phi}^{\wedge})$.
From the derivation in \cite{barfoot2014associating}, we can model the Gaussian uncertainty of rotation using the right perturbation as $\bm{R}=\bar{\bm{R}}\exp(\delta\bm{\phi}^{\wedge})\approx\bar{\bm{R}}(\bm{I}+\delta\bm{\phi}^{\wedge})$,
where $\bar{\bm{R}}$ is the noise-free rotation and $\delta\bm{\phi}\sim\mathcal{N}(\bm{0},\bm{\Sigma}_{\delta\bm{\phi}})$.
If $\bm{R}$ and $\bm{t}$ are considered simultaneously, we also use the 3D transformation matrix $\bm{T}\in SE(3)$ from the Lie group:
$\bm{T}=SE_{3}(\bm{R},\bm{t})=
  \begin{bmatrix}
    \bm{R} & \bm{t} \\
    \bm{0} & 1      \\
  \end{bmatrix}
$ to represent poses.

\begin{figure}[t]
  \centering
  \includegraphics[width=0.47\textwidth]{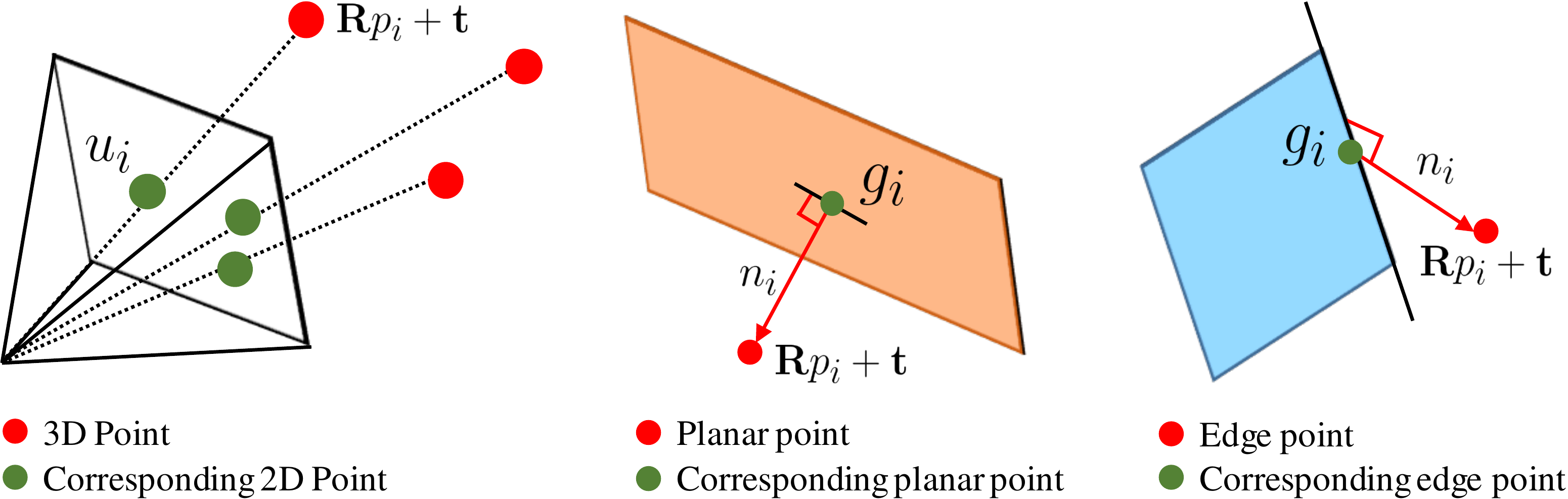}
  \caption{Visualization of geometric constraints of the PnP, PtPL, and PtL problems. The red dot is the reference point, while the green dot is the corresponding corner or edge feature.}
  \label{fig:qpep-geometric-constraint}
\end{figure}

\subsection{Quadratic Pose Estimation Problem}
\label{sec:ps_qpep}

As first proposed in our previous work \cite{wu2022quadratic}, QPEPs define a series of $SE(3)$-differentiable optimization problems as
\begin{equation}
  \underset{\bm{R}\in SO(3),\ \bm{t}\in\mathbb{R}^{3}}{\arg\min} \
  \mathcal{L}(\bm{R},\bm{t}),
\end{equation}
where $\mathcal{L}(\cdot)$ is represented as a \textit{quadratic objective} in terms of quadratic products of elements in $\bm{R}$ and $\bm{t}$. The PnP, PtPL, and PtL problems are three QPEPs since their objective are quadratic.
Fig. \ref{fig:qpep-geometric-constraint} visualizes the geometry of these problems.
Regarding the PnP problem, our goal is to estimate the transformation between the world frame and the camera frame given a set of 3D-2D point correspondences $<\bm{p}_{i}, \bm{u}_{i}>$:
\begin{equation}
  \label{equ:qpep-pnp}
  \mathcal{L}_{pnp}(\bm{R},\bm{t})
  =
  \sum||\bm{u}_{i} - \pi(\bm{R}\bm{p}_{i}+\bm{t})||^{2},
\end{equation}
where $\pi(\cdot)$ projects a 3D point onto the image plane.
The PtPL problem obtains the optimal transformation between two 3D point clouds by minimizing the point-to-plane residuals:
\begin{equation}
  \label{equ:qpep-ptpl}
  \mathcal{L}_{ptpl}(\bm{R},\bm{t})
  =
  \sum[\bm{n}_{i}^{\top}(\bm{g}_{i} - \bm{R}\bm{p}_{i} - \bm{t})]^{2},
\end{equation}
where $\bm{g}_{i}$ and $\bm{p}_{i}$ are $i$-th corresponding points from two point clouds and $\bm{n}^{i}$ is the $i$-th normal vector of a plane on which $\bm{g}_{i}$ stay.
For the PtL problem \cite{jiao2021robust}, we define the edge residual as a planar residual using \eqref{equ:qpep-ptpl},
where $\bm{n}_{i}$ coincides with the projection direction from the line to $\bm{p}_{i}$ (see Fig. \ref{fig:qpep-geometric-constraint}).
This formulation allows us to address both the PtPL and PtL problems with one solution.
Our previous work has proposed a unified quaternion-based globally optimal solution to solve general QPEPs \cite{wu2022quadratic}.
The mathematical derivation, solution strategies, and analysis of uncertainty are already detailed in this paper \cite{wu2022quadratic}.
In subsequent sections, we introduce how to extract features from raw sensor measurements and apply the \textit{QPEP-PnP} and \textit{QPEP-PtPL} algorithms to address the extrinsic calibration.

\section{Measurement Processing}
\label{sec:measure_process}

\begin{figure}[t]
  \centering
  \subfigure[]
  {\label{fig:mp_pipeline_result_cloud}\centering\includegraphics[width=0.19\textwidth]{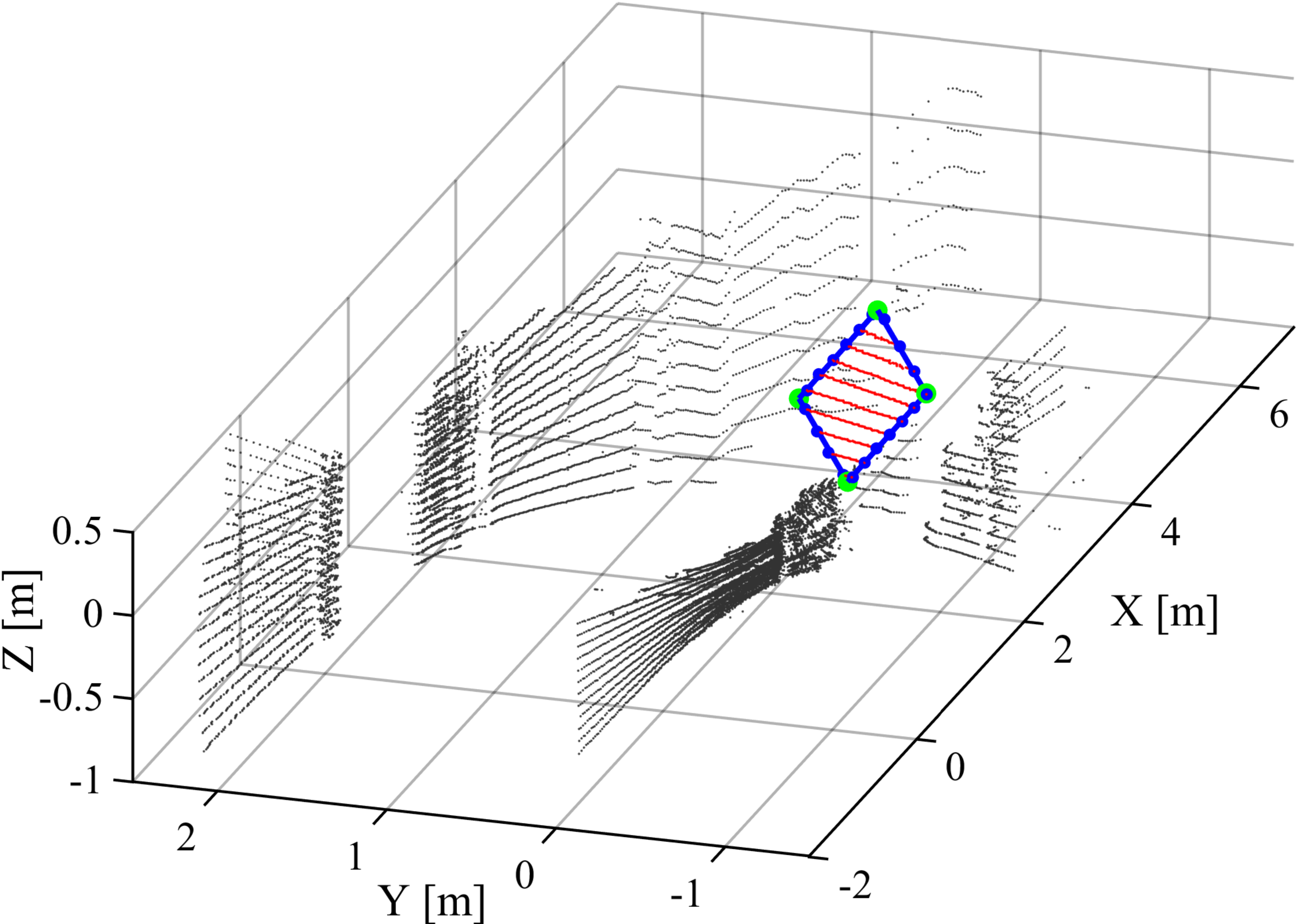}}
  \subfigure[]
  {\label{fig:mp_pipeline_result_frame_image}\centering\includegraphics[width=0.14\textwidth]{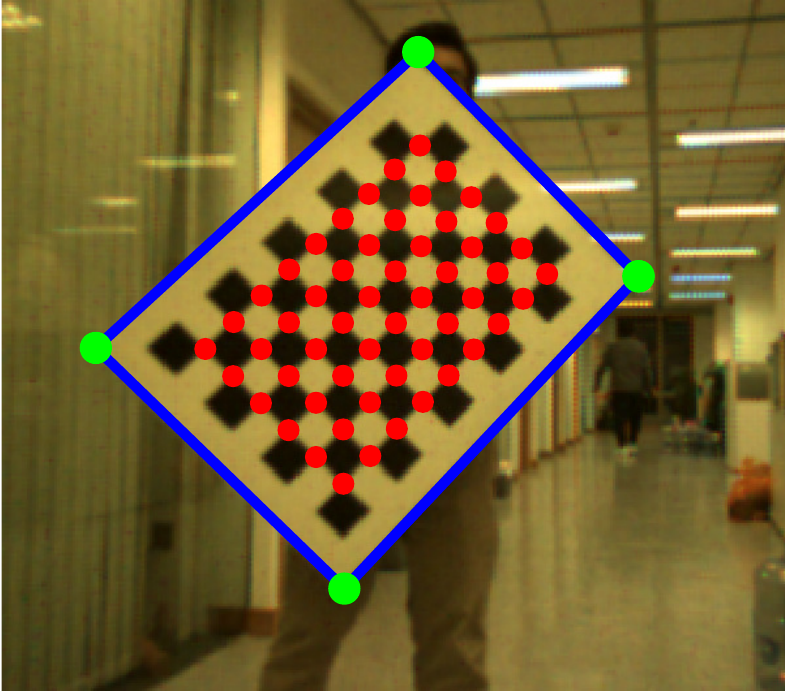}}
  \subfigure[]
  {\label{fig:mp_pipeline_result_event_image}\centering\includegraphics[width=0.14\textwidth]{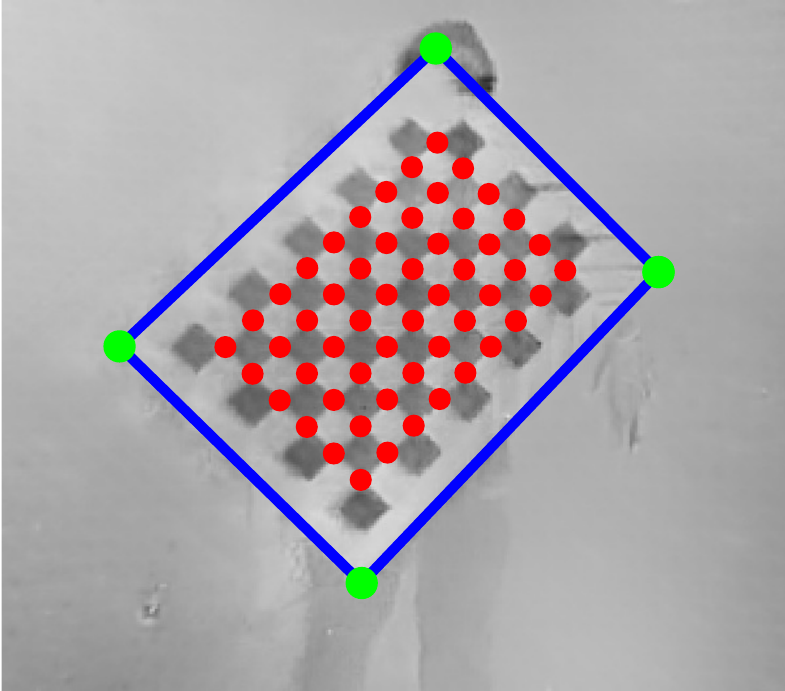}}
  \caption{Features of the checkerboard are extracted from (a) the point cloud, (b) frame image, and (c) reconstructed image from events.}
  \label{fig:mp_pipeline_result}
\end{figure}

This section explains how raw sensor data from LiDARs, frame cameras, and event cameras are preprocessed before estimating their extrinsics.
We can take advantage of the structural prior of the checkerboard for reliable feature selection and matching.
Fig. \ref{fig:mp_pipeline_result} shows the feature extraction results.

\subsection{Automatic Feature Extraction From LiDARs}
\label{sec:methodology_fsl}
We are interested in extracting three types of features from the checkerboard point cloud: edge points, planar points, and planar coefficients.
These features are useful in LiDAR-camera data association.
Since we work on mechanical scanning LiDARs, the following sections use \textit{ring} to denote the points set from the same emitter.
We propose a two-stage method for automatic feature extraction.

\subsubsection{Checkerboard Point Selection}
\label{sec:methodology_fsl_selection}
We need to select board points from the raw point cloud.
We have these observations to design the point selection method:
\textit{1)} the width and height of the board are known as a prior;
\textit{2)} points from the same ring form a straight line segment if they lie on the board;
\textit{3)} points from different rings form a planar patch if they lie on the board and stay near.

Firstly, we compute the pitch angle $\phi = \arctan(\frac{z}{\sqrt{x^2 + y^2}})$ for each point and split out points into different rings.
Some LiDARs directly provide the ``ring'' as a property of each point.
We then cluster each ring into multiple line segments based on the distance of two consecutive points.
The principle component of each line segment is computed by PCA: $[\sigma_1, \sigma_2, \sigma_3]$ where $\sigma_1 \geq \sigma_2 \geq \sigma_3$. The curvature is measured as $\sigma_1/\sigma_2$.
We keep line segments if their curvature is bigger than the predefined threshold $\mu_1$ and the length of the line segment smaller than an empirical threshold $\mu_2$.
Secondly, we utilize the DBScan algorithm \cite{kriegel2011density} to segment the remaining points as several clusters. Small clusters are then removed.
Thirdly, we verify each cluster by registering it with a board point template.
We compute the registration error for each cluster and keep the only one with the lowest error. The selected cluster is regarded as the candidate board points.

\subsubsection{Checkerboard Feature Extraction}
\label{sec:methodology_fsl_feature_extraction}

After obtaining board points, we use the RANSAC-based plane fitting method \cite{rusu20113d} to compute the normal vector $\bm{n}^{l}$, filter out outliers, and obtain a set of planar points $\mathcal{P}^{l}_{pl}$ (inliers).
Due to uncorrected bias and noise of LiDARs' measurements, we observe that board points form a $3cm$-thick plane approximately.
To reduce this effect in the subsequent calibration process, we project each planar point $\bm{p}$ onto the fitted plane as
\begin{equation}
  \label{equ:projection_on_plane}
  \bm{p}_{proj}^{l}
  =
  \bm{p}^{l} - a\bm{n}^{l},
  \ \ \
  a = \bm{n}^{l}\cdot(\bm{p}^{l} - \bm{g}^{l}),
\end{equation}
where $\bm{g}^{l}$ is a point on the plane.
The edge points $\mathcal{P}^{l}_{edge}$ are then extracted as the starting and ending points of each ring.

\subsubsection{Checkerboard Tracking}

Estimating the board's position allows us to track the board at the next frame.
The time-consuming point selection stage (costs $50-60ms$) is skipped.
We create a virtual bounding box around the checkerboard with a slightly larger size.
In the next frame, only points inside the box are kept.
This tracking stage also improves the success rate of the board point selection.

\subsection{Automatic Feature Extraction From Frame Cameras}
\label{sec:methodology_fsc}

The coefficients of the board observed by the frame camera, denoted by $[\bm{n}^{c},d^{c}]^{\top}$, can be easily obtained.
We use the off-the-shelf software to detect the inner patterns of the checkerboard automatically.
The board's pose $SE_{3}(\bm{R}^{c}_{b}, \bm{t}^{c}_{b})$ is estimated by the \textit{QPEP-PnP} algorithm, where we minimize the point-wise distance between 3D inner patterns and their corresponding 2D points.
Along the transformed boundaries of the board, we can generate many ``fake'' edge points $\mathcal{P}_{edge}^{c}$.
We can also obtain the covariance matrix $\bm{\Sigma}_{\bm{\delta\phi}}$ of the checkerboard's rotation from the \textit{QPEP-PnP} algorithm.
Based on the uncertainty representation of 6-DoF poses in Section \ref{sec:ps_notation}, we can propagate the uncertainty of the rotated normal vector $\bm{n}^{c}\approx\bar{\bm{R}}^{c}_{b}(\bm{I} + \delta\phi^{\wedge})\bm{n}^{b}$ as
\begin{equation}
  \begin{split}
    \bm{\Sigma}_{\bm{n}^{c}}
    \approx
    [\bar{\bm{R}}^{c}_{b}(\bm{n}^{b})^{\wedge}]
    \bm{\Sigma}_{\bm{\delta\phi}}
    [\bar{\bm{R}}^{c}_{b}(\bm{n}^{b})^{\wedge}]^{\top},
  \end{split}
\end{equation}
where the derived covariance will be used in the optimization objective in Section \ref{sec:ext_calibration}.


\subsection{Image Reconstruction From Events}
\label{sec:methodology_ire}
\subsubsection{Event Data}
An event camera has independent pixels that respond to logarithmic intensity change $L$.
In a noise-free scenario, an event $\bm{e}_{k}=[\bm{u}_{k}^{\top},t_{k},p_{k}]^{\top}$ is triggered at pixel $\bm{u}_{k} = [x_{k},y_{k}]^{\top}$ and time $t_{k}$ as soon as the logarithmic intensity increment reaches a contrast threshold $\pm C$ since the last event at the pixel, i.e.
\begin{equation}
  \Delta L(\bm{u}_{k},t_{k})
  \doteq
  L(\bm{u}_{k}, t_{k}) - L(\bm{u}_{k}, t_{k}-\Delta t_{k})
  \geqslant
  p_{k}C,
\end{equation}
where $C > 0$, $\Delta t_{k}$ is the time elapsed since the last event at the same pixel, and the polarity $p_{k}\in\{+1,-1\}$ is the sign of the intensity change \cite{gallego2020event}.

\begin{algorithm}[t]
  \caption{LiDAR-Camera Extinsic Calibration}
  \label{alg:ext_refinement}
  \LinesNumbered
  \KwIn{Number of iteration: $I$ \ \ \ \ \ \ \ \ \ \ \ \ \ \ \ \ \ \ \ \ \ \ \ \
    Number of LiDAR-camera data pairs: $N$
    Features: $\{\bm{n}^{c}\}$, $\{\mathcal{P}^{c}_{edge}\}$,
    $\{\mathcal{P}^{l}_{pl}\}$, $\{\mathcal{P}^{l}_{edge}\}$\\}
  \KwOut{Estimated extrinsics $\bm{T}^{c}_{l}$;}
  Initialize extrinsics $\bm{T}^{c}_{l,ini}=
    \textit{QPEP-PtPL}(\{\bm{n}^{c}\},\{\mathcal{P}^{l}_{pl}\})$\\
  \ForEach{$M\in\{1,2, \dots, N\}$}
  {
  The set of all candidate extrinsics: $\mathcal{T}=\emptyset$\\
  \While{count of iteration $<I$}
  {
  Randomly select $M$ data pairs: $\{\bm{n}^{c}\}^{\#}$, $\{\mathcal{P}^{c}_{edge}\}^{\#}$,
  $\{\mathcal{P}^{l}_{pl}\}^{\#}$, $\{\mathcal{P}^{l}_{edge}\}^{\#}$\\
  Find corresponding edge points of $\{\mathcal{P}^{l}_{edge}\}^{\#}$:
  $\{\mathcal{E}^{c}\}\subseteq\{\mathcal{P}^{c}_{edge}\}^{\#}$ given $\bm{T}^{c}_{l,ini}$\\
  Optimize extrinsics: $\bm{T}=$
  $\textit{QPEP-PtPL}(\{\bm{n}^{c}\}^{\#},$
    $\{\mathcal{P}^{l}_{pl}\}^{\#},\{\mathcal{E}^{c}\}, \{\mathcal{P}^{l}_{edge}\}^{\#})$\\
  $\mathcal{T}=\mathcal{T}\cup\bm{T}$\\
  }
  Compute the mean of $\mathcal{T}$: $\bar{\bm{T}}$\\
  $\bm{T}^{c}_{l}=\bar{\bm{T}}$ if the mean geometric error is smaller\\
  }
\end{algorithm}

\begin{figure}[t]
  \centering
  \subfigure[]
  {\label{fig:exp_rlfs01_cloud_align}\centering\includegraphics[width=0.257\textwidth]{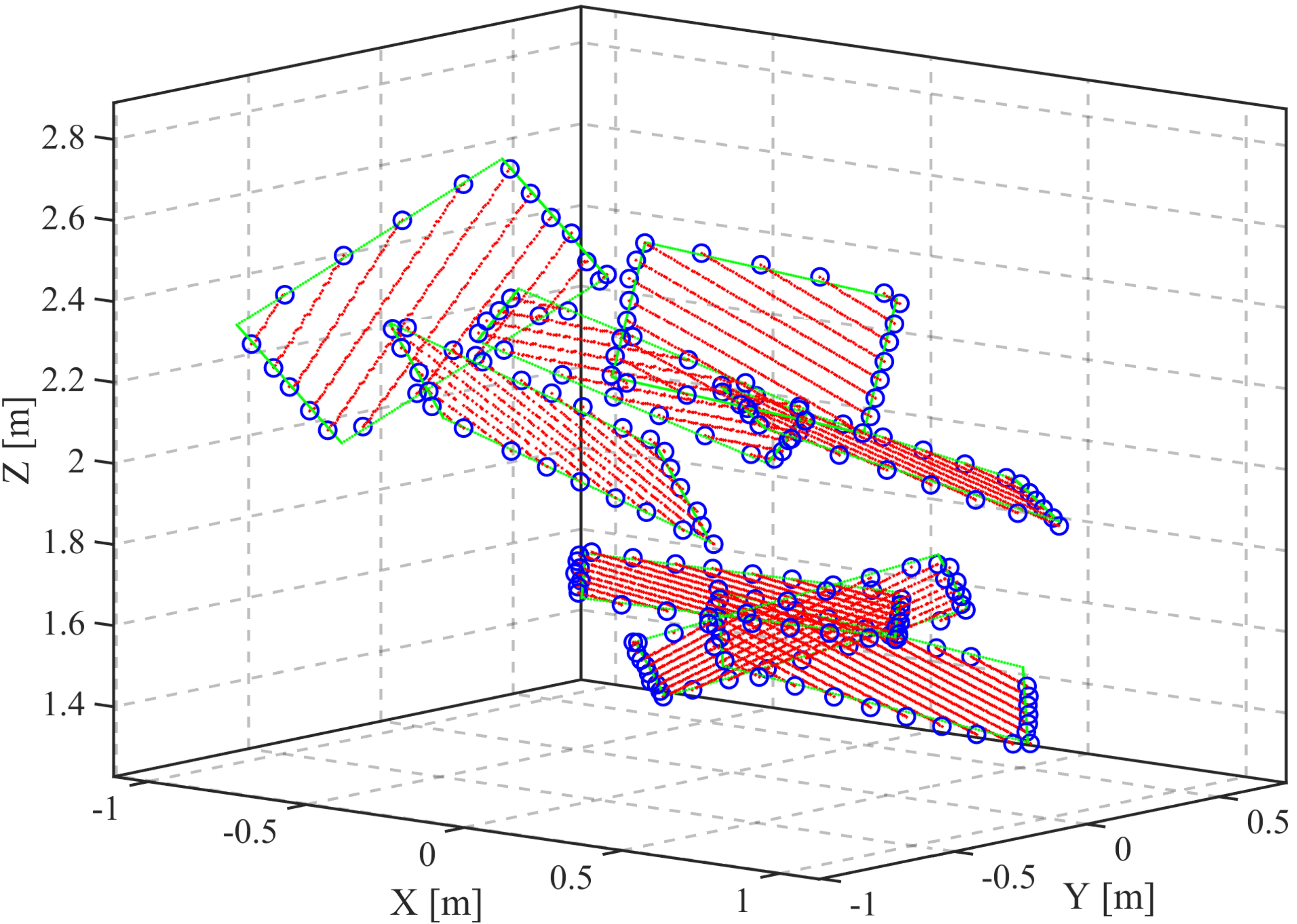}}
  \subfigure[]
  {\label{fig:exp_rles03_cloud_align}\centering\includegraphics[width=0.22\textwidth]{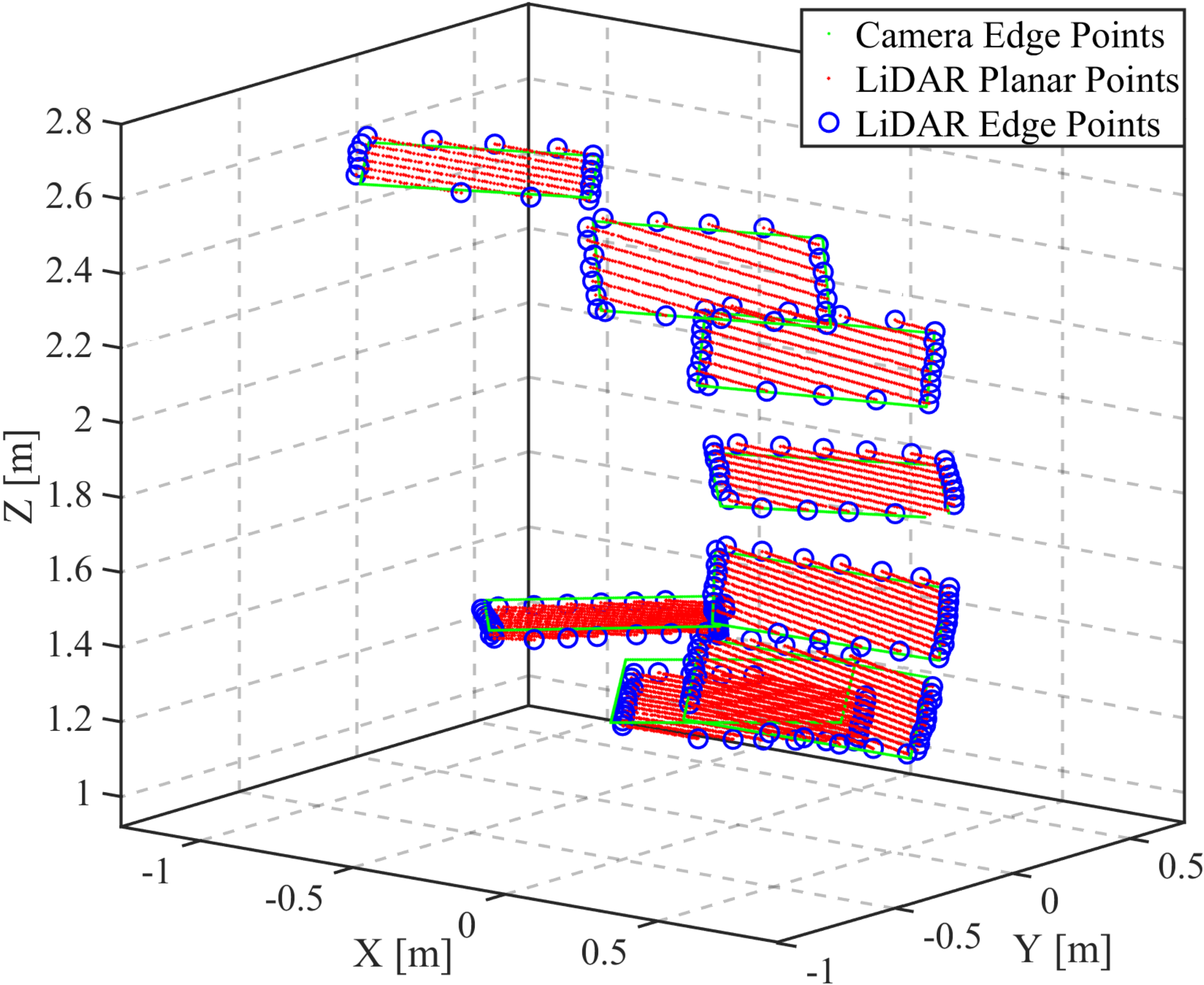}}
  \caption{The LiDAR's planar points and edge points are aligned with the board plane of images with the estimated extrinsiscs.}
  \label{fig:exp_cloud_align}
\end{figure}

\subsubsection{Image Reconstruction}
One of the fundamental building blocks for camera calibration is the detection of checkerboard corners.
However, these corner detectors originally designed for frame images are not directly applicable to events due to their intrinsically asynchronous and sparse nature.
Inspired by the work done by Muglikar \textit{et al.} \cite{muglikar2021calibrate}, we resort to a learning-based method called E2VID \cite{rebecq2019high} to reconstruct high-quality frame images from the asynchronous and sparse event stream.
This method encodes events in a spatio-temporal voxel grid and uses a recurrent convolutional neural network based on the UNet architecture \cite{ronneberger2015u} to process events.
The network is already pretrained using a large number of simulated event sequences.
After the reconstruction, we can directly apply the approach in Section \ref{sec:methodology_fsc} on these images for calibration.

The reconstruction procedure is summarized in three steps:
1) Move the checkerboard before the event camera to trigger events.
2) Divide events into chunks of constant time duration ($50ms$ in our experiments). But the time duration of these chunks does not have to be constant. One could choose to define the chunks by the number of events.
3) Reconstruct images from events using E2VID. The corresponding LiDAR frame specifies the reconstruction timestamp.

Besides the above approach, we also have two possible solutions to detect corners from events, but they present several limitations.
The first solution is to aggregate events within a local spatial-temporal window to create an event map that is also an image-type representation \cite{jiao2021comparing}.
But event maps do not record intensity and contain several noisy pixels, making the traditional corner detectors inaccurate.
Another approach is to detect corners from pure events.
However, existing event-based corner detectors \cite{manderscheid2019speed} are not specifically designed for checkerboard corners, and most of them are not publicly released, inducing difficulties in calibration.
In contrast, the proposed reconstruction-based calibration scheme is easy to implement and does not require much parameter tuning.
Experimental results have demonstrated the effectiveness of the proposed reconstruction-based calibration method.

\section{Extrinsic Calibration with Multi-Frame Measurements}
\label{sec:ext_calibration}

\begin{figure}[t]
  \centering
  \subfigure[LiDAR-Frame camera sensor]
  {\label{fig:exp_real_device_1}\centering\includegraphics[width=0.21\textwidth]{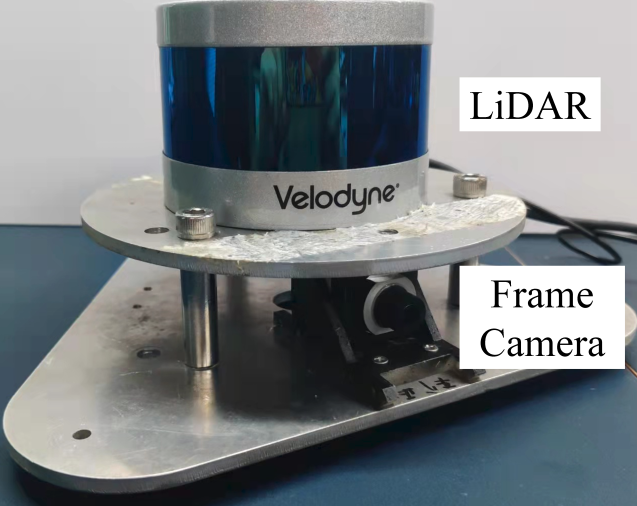}}
  \subfigure[LiDAR-Event camera sensor]
  {\label{fig:exp_real_device_2}\centering\includegraphics[width=0.219\textwidth]{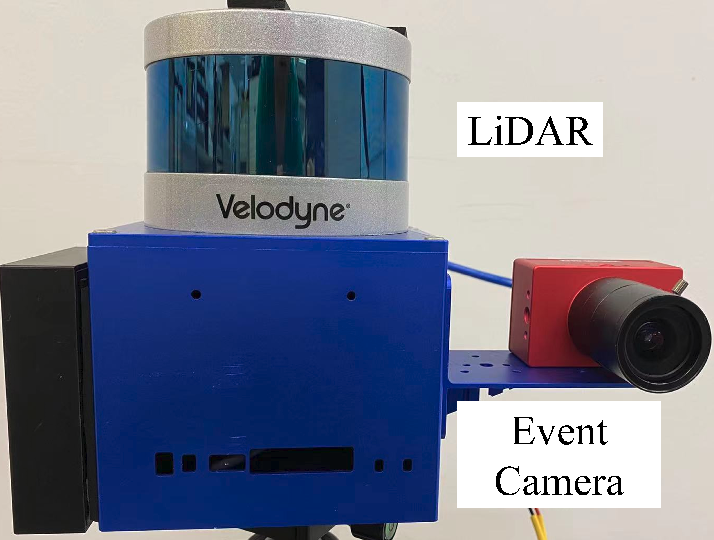}}
  \caption{The RLFS and RLES devices for calibration tests.}
  \label{fig:exp_real_device}
\end{figure}

\begin{figure}[t]
  \centering
  \subfigure[]
  {\label{fig:exp_simu_camera}\centering\includegraphics[width=0.210\textwidth]{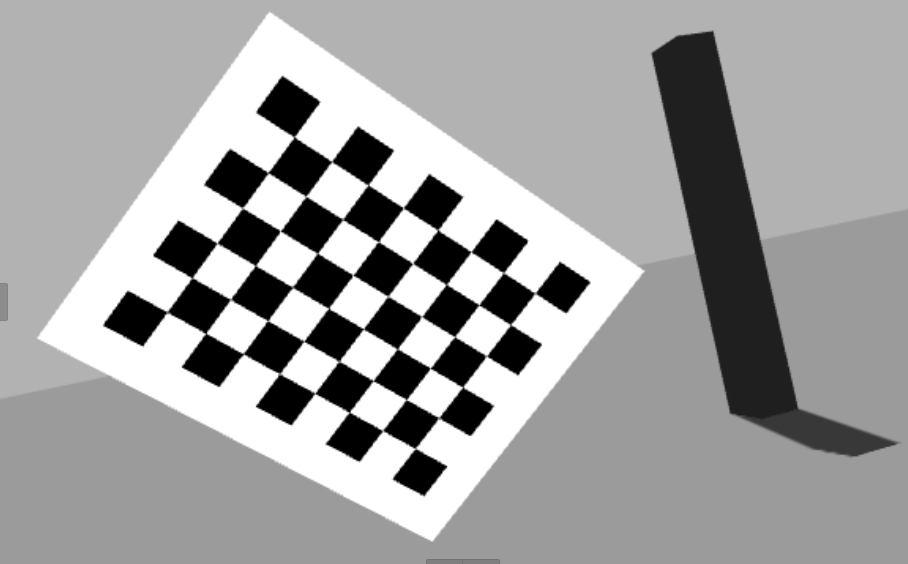}}
  \
  \subfigure[]
  {\label{fig:exp_simu_lidar}\centering\includegraphics[width=0.21\textwidth]{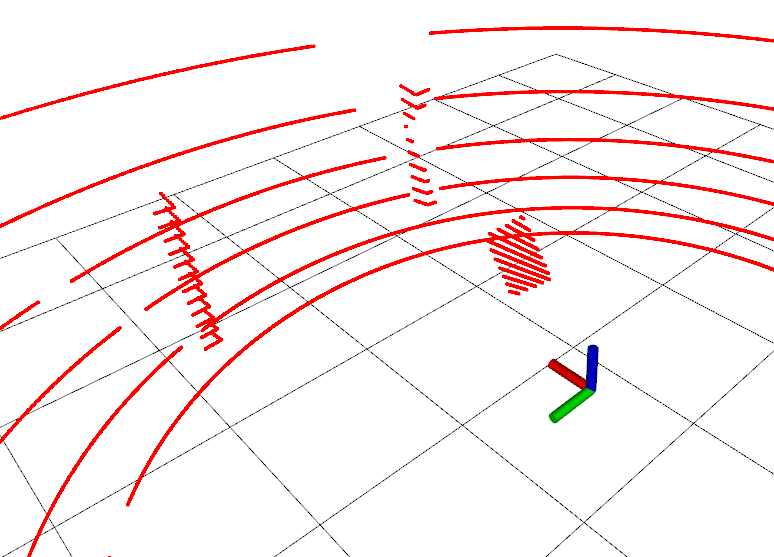}}
  \caption{Simulated sensor data: (a) an image and (b) a point cloud.}
  \label{fig:exp_simu_sensor}
\end{figure}

Section \ref{sec:measure_process} has explained how checkerboard's normal vectors: $\bm{n}^{c}$, $\bm{n}^{e}$, planar points: $\mathcal{P}^{l}_{pl}$, and edge points: $\mathcal{P}^{l}_{edge}$, $\mathcal{P}^{c}_{edge}$, $\mathcal{P}^{e}_{edge}$ are extracted from each frame of LiDARs, frame cameras, and event cameras.
This section introduces how to use these features to constrain extrinsics.
We take the LiDAR-frame camera extrinsic calibration as an example without loss of generality.
At first, we can easily construct a set of point-plane pairs.
But the point-edge association is ambiguous without the initial extrinsics.
In other words, we cannot determine which boundaries a point belongs to since the board's shape is symmetric.
Thus, we follow the \textit{initialization-refinement} philosophy \cite{jiao2021robust} to estimate extrinsics in a coarse-fine manner, as summarized in Algorithm \ref{alg:ext_refinement}.

The first phase initializes the extrinsics $\bm{T}^{c}_{l,ini}$
(i.e., transformation from $()^{c}$ to $()^{l}$).
We formulate the point-to-plane optimization objective \eqref{equ:qpep-ptpl}
by utilizing planar features: $\bm{n}^{c}$ and $\mathcal{P}^{l}_{pl}$
from several frames of measurements.
The \textit{QPEP-PtPL} algorithm is applied to solve the problem.

The second phase further refines the initial solution by jointly utilizing the planar and edge points that offer stronger geometric constraints.
This phase is executed with multiple computations.
Specifically, three steps are done at line $6$-$7$:
\begin{enumerate}
  \item With $\bm{T}^{c}_{l,ini}$, we first transform $\mathcal{P}^{l}_{edge}$ into the camera frame. For each point from transformed point set, we find the nearest point in $\mathcal{P}^{c}_{edge}$ as its corresponding edge point, and then construct the point-to-line objective.
  \item From \eqref{equ:qpep-ptpl}, we propagate the covariance of each point-to-plane distance as
        $\sigma^{2}=(\bm{R}^{c}_{l}\bm{p}_{i}^{l}+\bm{t}^{c}_{l})^{\top}\bm{\Sigma}_{n^{c}}(\bm{R}^{c}_{l}\bm{p}_{i}^{l}+\bm{t}^{c}_{l})$. We filter out $10\%$ error terms if their variances are large.
  \item The overall optimization objective for the refinement is defined as the sum of all weighted point-to-plane and point-to-line residuals as
        \begin{equation}
          \label{equ:obj-refinement}
          \mathcal{L}(\bm{R}^{c}_{l},\bm{t}^{c}_{l})
          =
          w_{ptpl}\mathcal{L}_{ptpl}(\bm{R}^{c}_{l},\bm{t}^{c}_{l})
          +
          w_{ptl}\mathcal{L}_{ptl}(\bm{R}^{c}_{l},\bm{t}^{c}_{l}),
        \end{equation}
        where $w_{ptpl}$ and $w_{ptl}$ are tunned parameters which balance these two constraints.
        Similar to initialization, we can also use the \textit{QPEP-PtPL} solver.
\end{enumerate}

After getting a set of candidate extrinsics, we select one with the minimum geometric error as our resulting extrinsics.
Two examples are shown in Fig. \ref{fig:exp_cloud_align}, where LiDAR's planar points and edge points are aligned with the board plane of images with the estimated extrinsiscs.

\section{Experiment}
\label{sec:experiment}




\begin{figure}[t]
  \centering\includegraphics[width=0.45\textwidth]{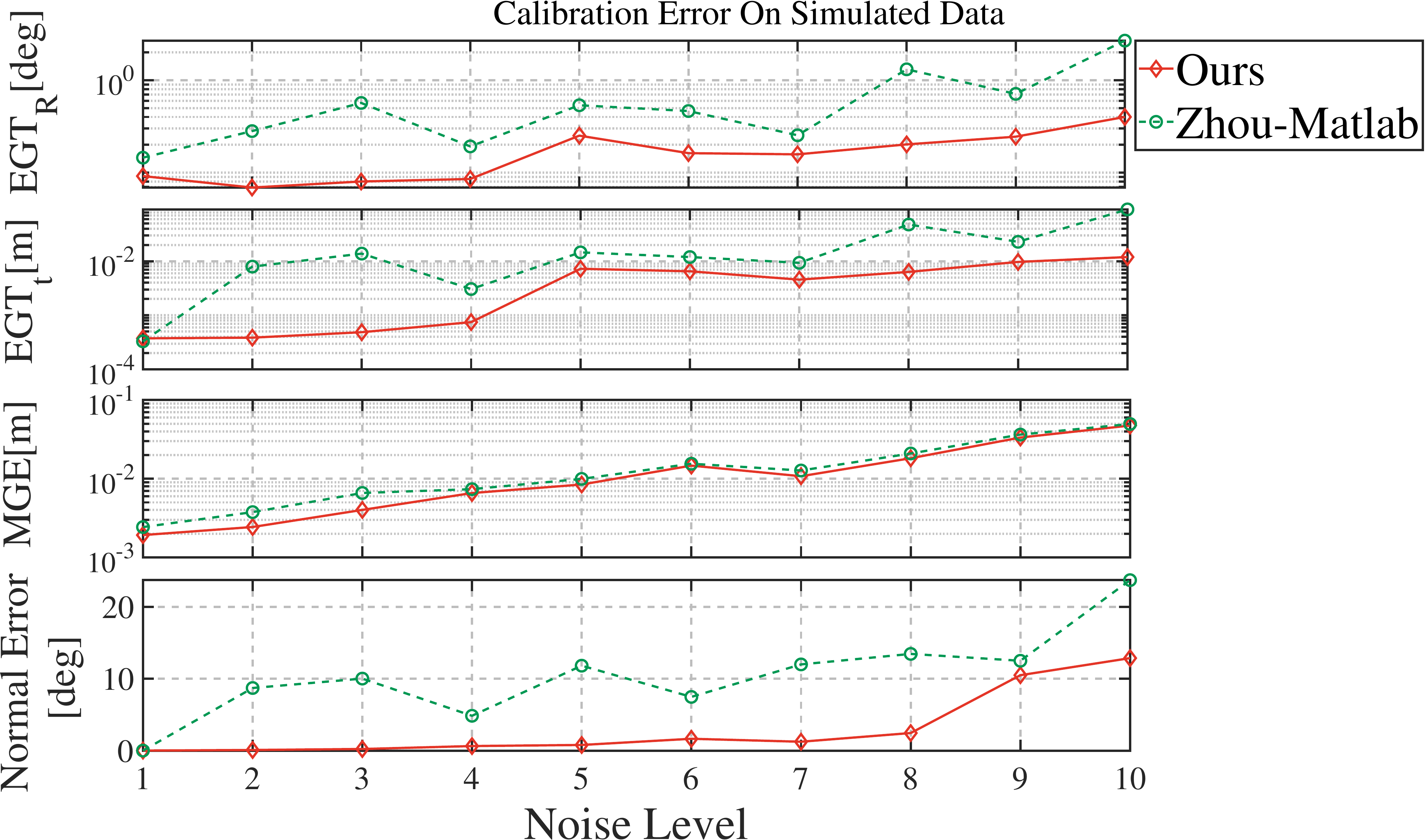}
  \caption{Calibration error of the proposed method and \textit{Zhou-MATLAB} on calibration data at different levels of noise. The proposed method can achieve calibration accracy (the top three rows) ranging from $[0.068,0.400]deg$, $[0.0004, 0.012]m$, $[0.0003, 0.024]m$, $[0.002, 0.023]m$ at    $EGT_{\bm{R}}$, $EGT_{\bm{t}}$, and $MGE$ respectively. Our method achieves the plane fitting accuracy (the fourth row) that is computed as the relative angle with the ground-truth normal ranging from $[0.0104,12.8701]deg$.}
  \label{fig:exp_err_plot_simu_data}
\end{figure}

\begin{figure*}[t]
  \centering
  \subfigure[Median rotation error]
  {\label{fig:exp_rlfs_error_rot}\centering\includegraphics[width=0.232\textheight]{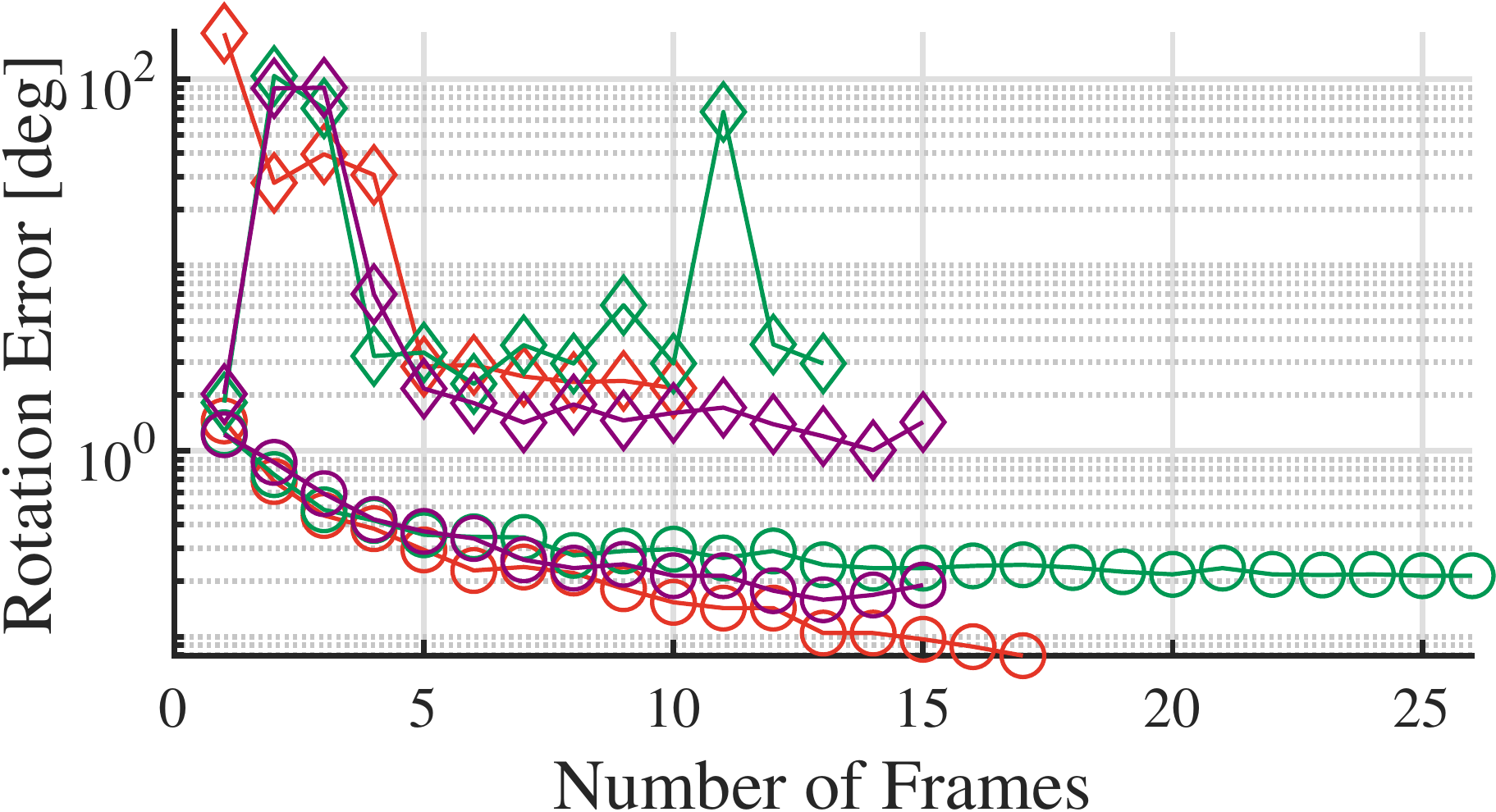}}
  \subfigure[Median translation error]
  {\label{fig:exp_rlfs_error_tsl}\centering\includegraphics[width=0.232\textheight]{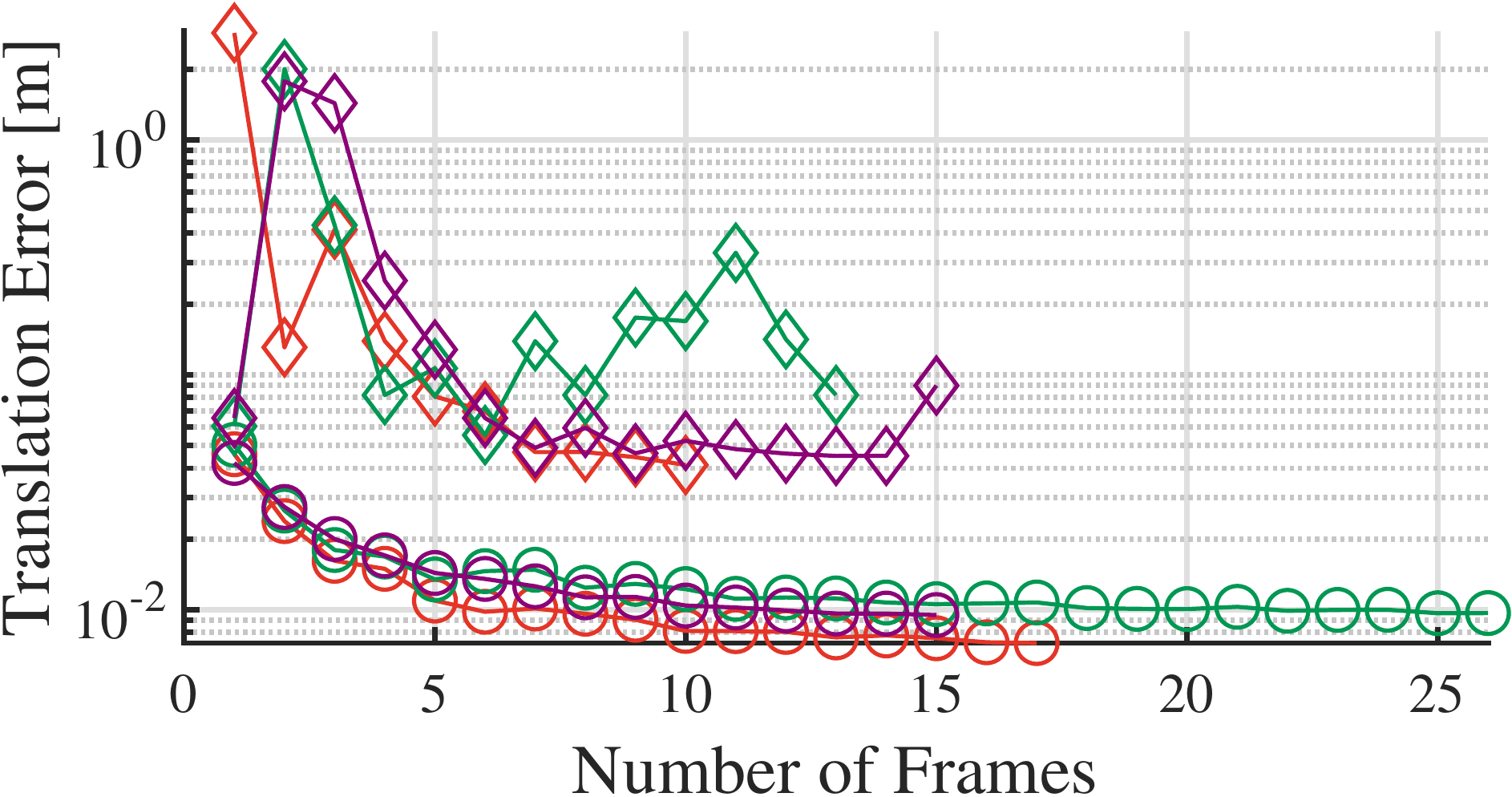}}
  \subfigure[Median MGE]
  {\label{fig:exp_rlfs_error_mge}\centering\includegraphics[width=0.263\textheight]{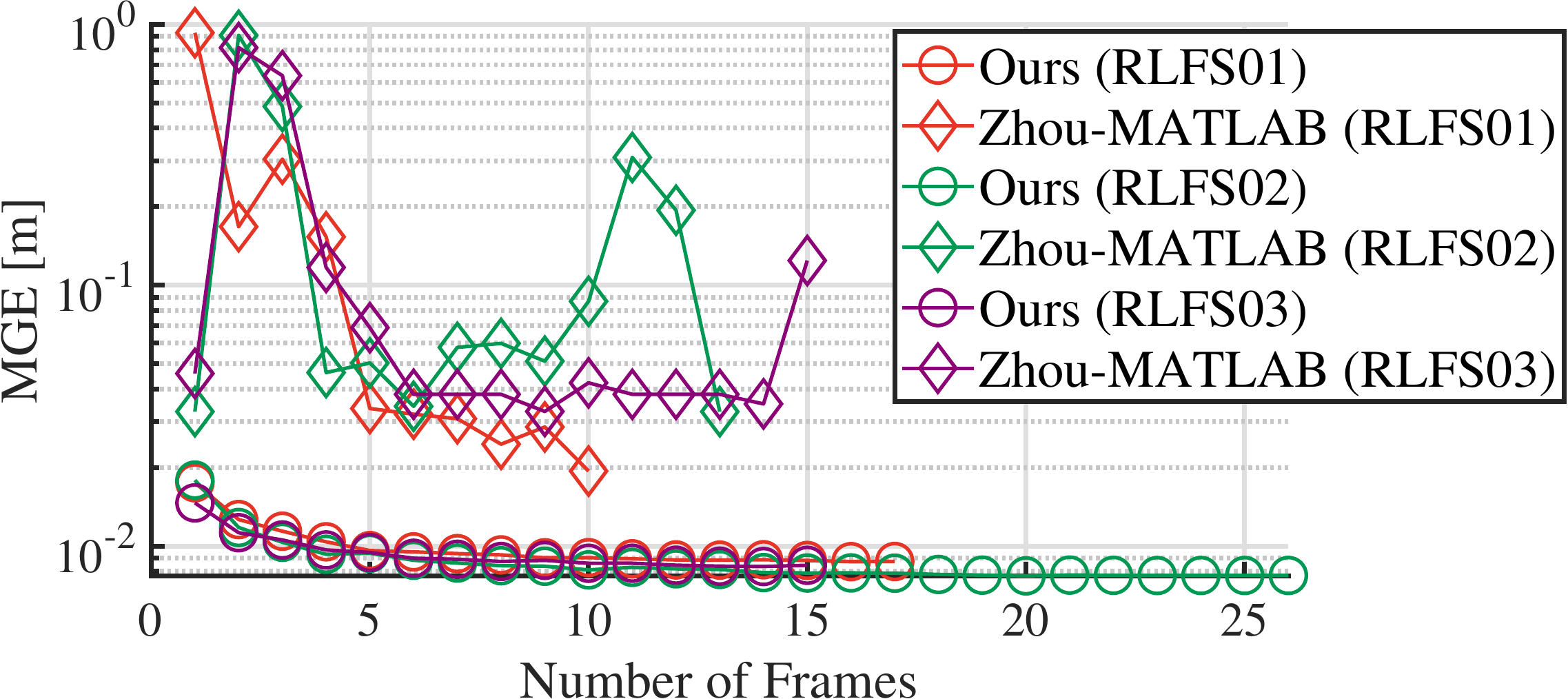}}
  \caption{Median rotation error, translation error, and MGE of ours and \textit{Zhou-MATLAB} on calibration data: \textit{RLFS01}-\textit{RLFS03}. The median errors are computed using the set of candidate extrinsics at line $8$ of Algorithm \ref{alg:ext_refinement}.}
  \label{fig:exp_rlfs_error}
\end{figure*}

\begin{table*}[]
  \centering
  \caption{Calibration results on \textit{RLFS}.
    $\downarrow$/$\uparrow$ indicates that the lower/higher the value, the better the score.}
  \renewcommand\arraystretch{0.9}
  \renewcommand\tabcolsep{4.6pt}
  \scriptsize
  \begin{tabular}{ccccccccccccc}
    \toprule
    \multirow{2}{*}{Dataset}                              &
    \multirow{2}{*}{Method}                               &
    \multicolumn{4}{c}{Quaternion}                        &
    \multicolumn{3}{c}{Translation {$[m]$}}               &
    \multirow{2}{*}{$EGT_{\mathbf{R}}\ [deg,\downarrow]$} &
    \multirow{2}{*}{$EGT_{\mathbf{t}}\ [m,\downarrow]$}   &
    \multirow{2}{*}{\textit{MGE}$\ [m,\downarrow]$}       &
    \multirow{2}{*}{$N_{\text{detect}}/N\ [\uparrow]$}                                                                                                                                                                                                                                                     \\
    \cline{3-9}
                                                          &                         & \multicolumn{1}{c}{$w$} & \multicolumn{1}{c}{$x$} & \multicolumn{1}{c}{$y$} & \multicolumn{1}{c}{$z$}
                                                          & \multicolumn{1}{c}{$x$} & \multicolumn{1}{c}{$y$} & \multicolumn{1}{c}{$z$} &                         &                         &                &                                                                                             \\
    \midrule[0.03cm]

                                                          & \textit{CAD}            & $\dgt{0.664}$           & $\dgt{0.664}$           & $\dgt{-0.242}$          & $\dgt{0.242}$           & $\dgt{0.000}$  & $\dgt{-0.059}$ & $\dgt{-0.070}$
                                                          & $\dgt{-}$               & $\dgt{-}$               & $\dgt{-}$               & $\dgt{-}$                                                                                                                                                        \\
                                                          & \textit{Fake GT}        & $\dgt{0.666}$           & $\dgt{0.667}$           & $\dgt{-0.233}$          & $\dgt{0.238}$           & $\dgt{-0.008}$ & $\dgt{-0.069}$ & $\dgt{-0.084}$
                                                          & $\dgt{-}$               & $\dgt{-}$               & $\dgt{-}$               & $\dgt{-}$                                                                                                                                                        \\
    \midrule

    \multirow{2}{*}{\textit{RLFS01}}
                                                          & \textit{Ours}           & $0.667$                 & $0.666$                 & $-0.233$                & $0.239$                 & $-0.003$       & $-0.074$       & $-0.091$       & $\bm{0.130}$ & $\bm{0.009}$ & $\bm{0.009}$ & $\bm{17}/18$ \\
                                                          & \textit{Zhou-MATLAB}    & $0.669$                 & $0.662$                 & $-0.238$                & $0.240$                 & $0.016$        & $-0.071$       & $-0.093$       & $0.850$      & $0.026$      & $0.011$      & $10/18$      \\
    \midrule

    \multirow{2}{*}{\textit{RLFS02}}
                                                          & \textit{Ours}           & $0.665$                 & $0.668$                 & $-0.234$                & $0.238$                 & $-0.000$       & $-0.060$       & $-0.090$       & $\bm{0.293}$ & $\bm{0.014}$ & $\bm{0.007}$ & $\bm{26}/29$ \\
                                                          & \textit{Zhou-MATLAB}    & $0.670$                 & $0.663$                 & $-0.227$                & $0.244$                 & $-0.008$       & $-0.110$       & $-0.095$       &
    $1.223$                                               & $0.042$                 & $0.013$                 & $21/29$                                                                                                                                                                                    \\
    \midrule

    \multirow{2}{*}{\textit{RLFS03}}
                                                          & \textit{Ours}           & $0.667$                 & $0.667$                 & $-0.233$                & $0.237$                 & $-0.010$       & $-0.068$       & $-0.092$       & $\bm{0.136}$ & $\bm{0.008}$ & $\bm{0.008}$ & $\bm{15}/15$ \\
                                                          & \textit{Zhou-MATLAB}    & $0.669$                 & $0.661$                 & $-0.238$                & $0.244$                 & $0.029$        & $-0.077$       & $-0.100$       & $1.120$      & $0.041$      & $0.013$      & $13/15$      \\

    \bottomrule[0.03cm]
  \end{tabular}
  \label{tab:rlfs_calibration_results}
\end{table*}

\begin{figure*}[t]
  \centering
  \subfigure[\textit{RLFS01}]
  {\label{fig:exp_rlfs_calibration_image_01}\centering\includegraphics[width=0.243\textwidth]{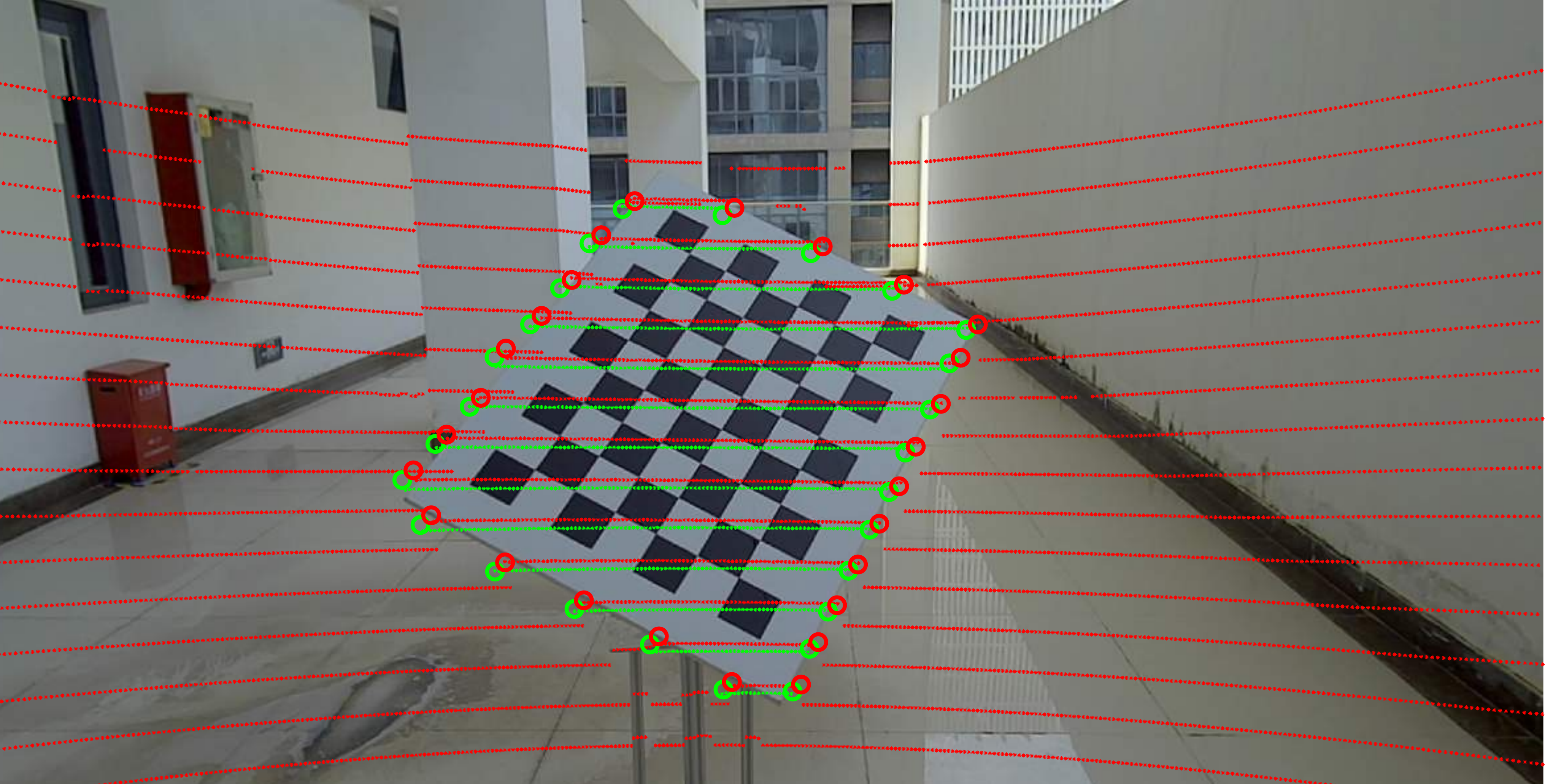}}
  \subfigure[\textit{RLFS02}]
  {\label{fig:exp_rlfs_calibration_image_02}\centering\includegraphics[width=0.241\textwidth]{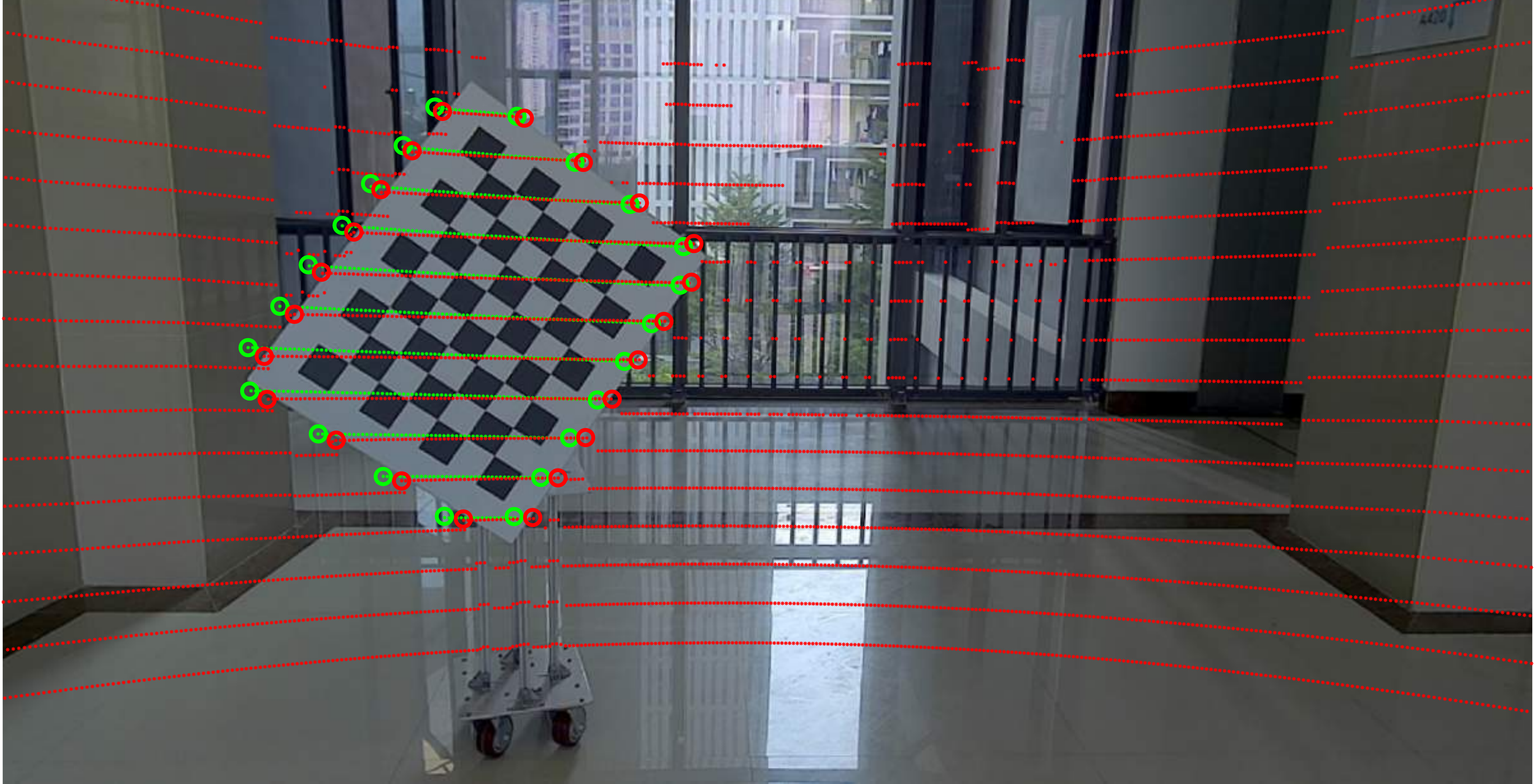}}
  \subfigure[\textit{RLFS03}]
  {\label{fig:exp_rlfs_calibration_image_03}\centering\includegraphics[width=0.220\textwidth]{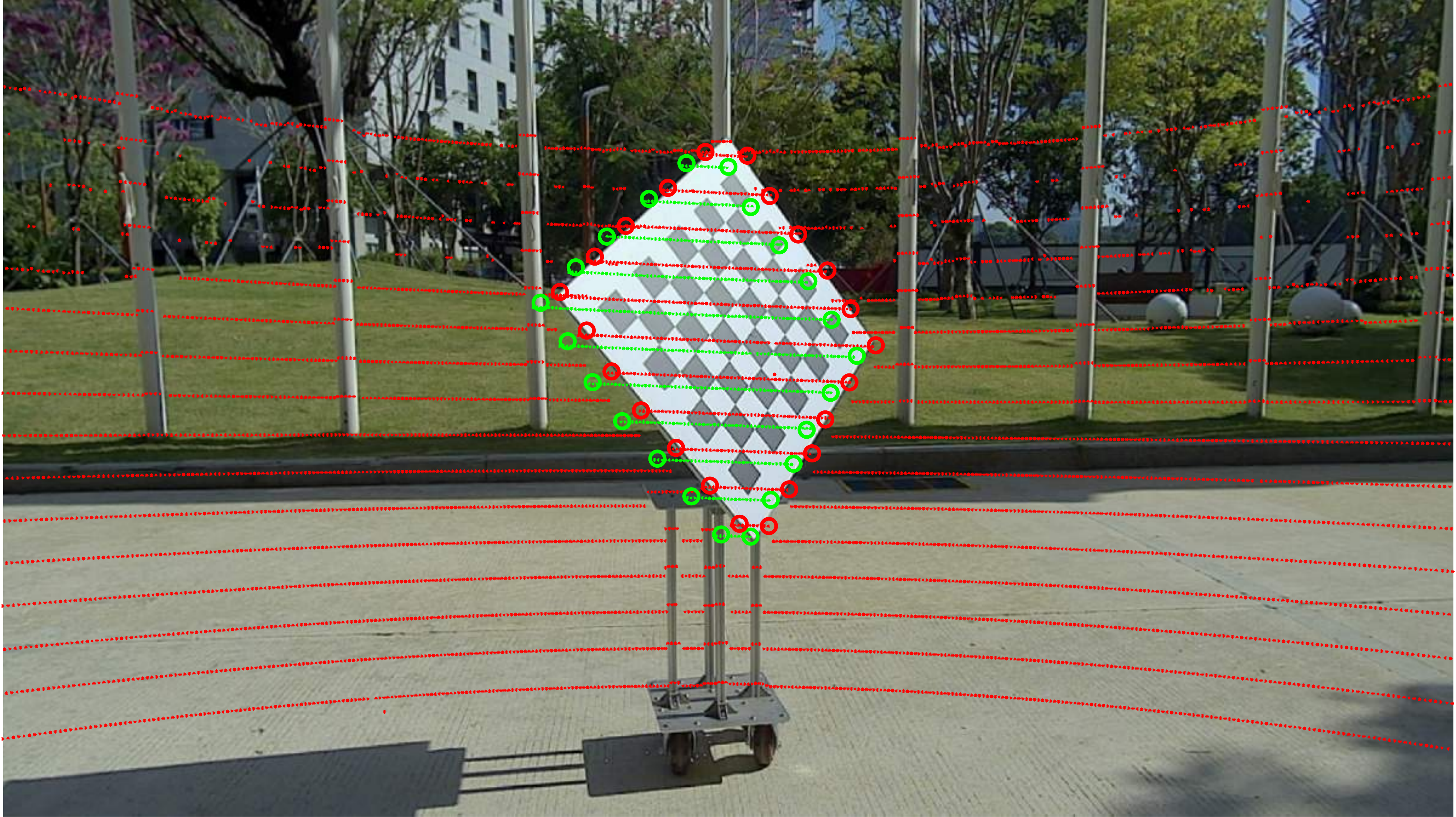}}
  \subfigure[\textit{RLFS02} (cloud)]
  {\label{fig:exp_rlfs_calibration_cloud_02_cloud}\centering\includegraphics[width=0.260\textwidth]{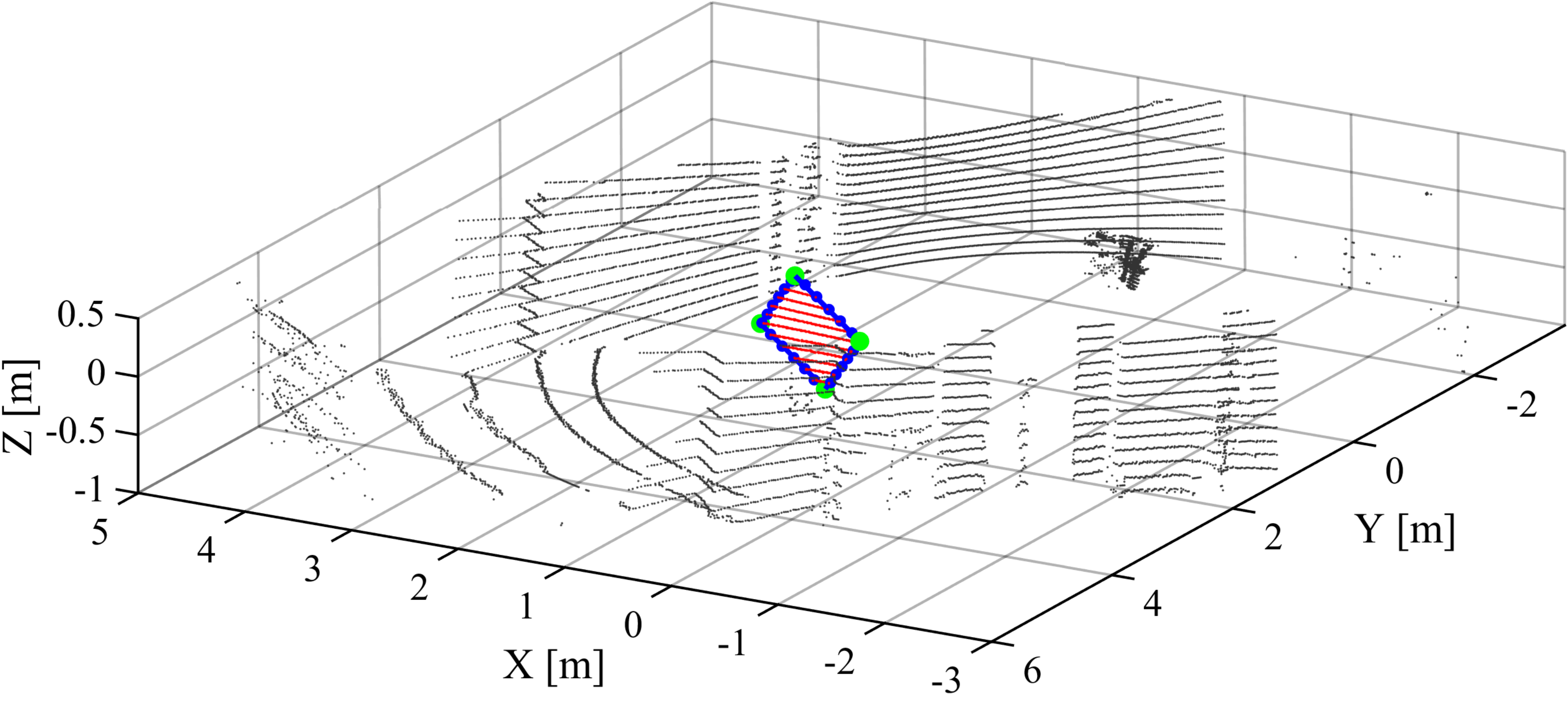}}
  \caption{Back-projection of the LiDAR points om images using the extrinsics from Table \ref{tab:rlfs_calibration_results} from our algorithm (red) and \textit{Zhou-MATLAB} (green).}
  \label{fig:exp_rlfs_calibration_image}
\end{figure*}

\subsection{Implementation Details}
\label{sec:exp_implementation}

We implement the proposed LCE-Calib in MATLAB.
We empirically set $\mu_1=0.9$ and $\mu_2=1$ in the checkboard point selection,
and $I=100$ in Algorithm \ref{alg:ext_refinement}.
Hyperparameters of the objective \eqref{equ:obj-refinement}: $w_{ptpl}$ and $w_{ptl}$, should be set according to a specific sensor configuration,
affected by the LiDAR's density and camera's resolution.

\subsubsection{Sensor Suites}
The proposed method is tested on both the simulated and real-world sensor suite.
The LiDAR and camera on real platforms are hardware-synchronized.
Hyperparameters: the number of checkerboard's inner patterns and board size should be set for a specific calibration target.
\begin{enumerate}
  \item \textit{The Simulated Sensor Suite} (SSS) is built on the Gazebo \cite{koenig2004design}. It consists of a simulated 16-beam LiDAR and frame camera with the same specification to RLFS. The ground-truth intrinsics and extrinsics are provided.

  \item \textit{The Real-World LiDAR-Frame Camera Sensor} (RLFS) contains a VLP-16 LiDAR\footnote{\url{https://velodynelidar.com/products/puck}} and a high-resolution camera (GSML-AR0143-H090, resolution: $1280$H$\times 720$V)\footnote{\url{https://docs.miivii.com/product/apex/xavier/manual/en/common/05.EN_gmsl.html}}, as shown in Fig. \ref{fig:exp_real_device_1}.

  \item \textit{The Real-World LiDAR-Event Camera Sensor} (RLES) is made for the mapping application and shown in Fig. \ref{fig:exp_real_device_2}. It is installed with a VLP-16 LiDAR and an event camera (DAVIS346, resolution: $360$H$\times 240$V)\footnote{\url{https://inivation.com/wp-content/uploads/2019/08/DAVIS346.pdf}}.
\end{enumerate}

Both frame and event cameras use the pinhole projection model.
And the DAVIS346 event camera outputs RGB frames that are used to verify the frame camera-based calibration.

\subsubsection{Evaluation Metrics}
We introduce two metrics to assess the LiDAR-camera calibration results from different aspects.
\begin{enumerate}
  \item \textit{Error Compared With Ground Truth (EGT)} computes the distance between the ground truth and the estimated values in terms of rotation and translation as
        \begin{equation}
          \begin{aligned}
            EGT_{\bm{R}}
             & =
            ||\ln(\bm{R}_{gt}\bm{R}_{est}^{-1})^{\vee}||, \\
            EGT_{\bm{t}}
             & =
            ||\bm{t}_{gt}-\bm{t}_{est}||.
          \end{aligned}
        \end{equation}
  \item \textit{Mean Geometric Error (MGE)} is the sum of the mean planar error (MPE) and mean edge error (MEE). MPE and MEE compute the mean point-to-plane and point-to-line distance between the 3D checkerboard plane in the camera frame and checkerboard points in the LiDAR frame respectively.
\end{enumerate}

The perfect ground truth is unknown in real-world experiments.
We first estimate 3D corner points from both images and point clouds.
We then manually check and modify them.
With 3D-3D point correspondences that are found, we can compute the ``fake'' ground truth with the ICP algorithm.
Our proposed method is compared with a baseline method (denoted by \textit{Zhou-MATLAB}) that has been released in MATLAB\footnote{\url{https://www.mathworks.com/help/lidar/ug/lidar-and-camera-calibration.html}},
This method implements the SOTA LiDAR-camera calibration method \cite{zhou2018automatic} but formulates a different objective function.

\begin{figure*}[t]
  \centering
  \subfigure[Median rotation error]
  {\label{fig:exp_rles_error_rot}\centering\includegraphics[width=0.31\textwidth]{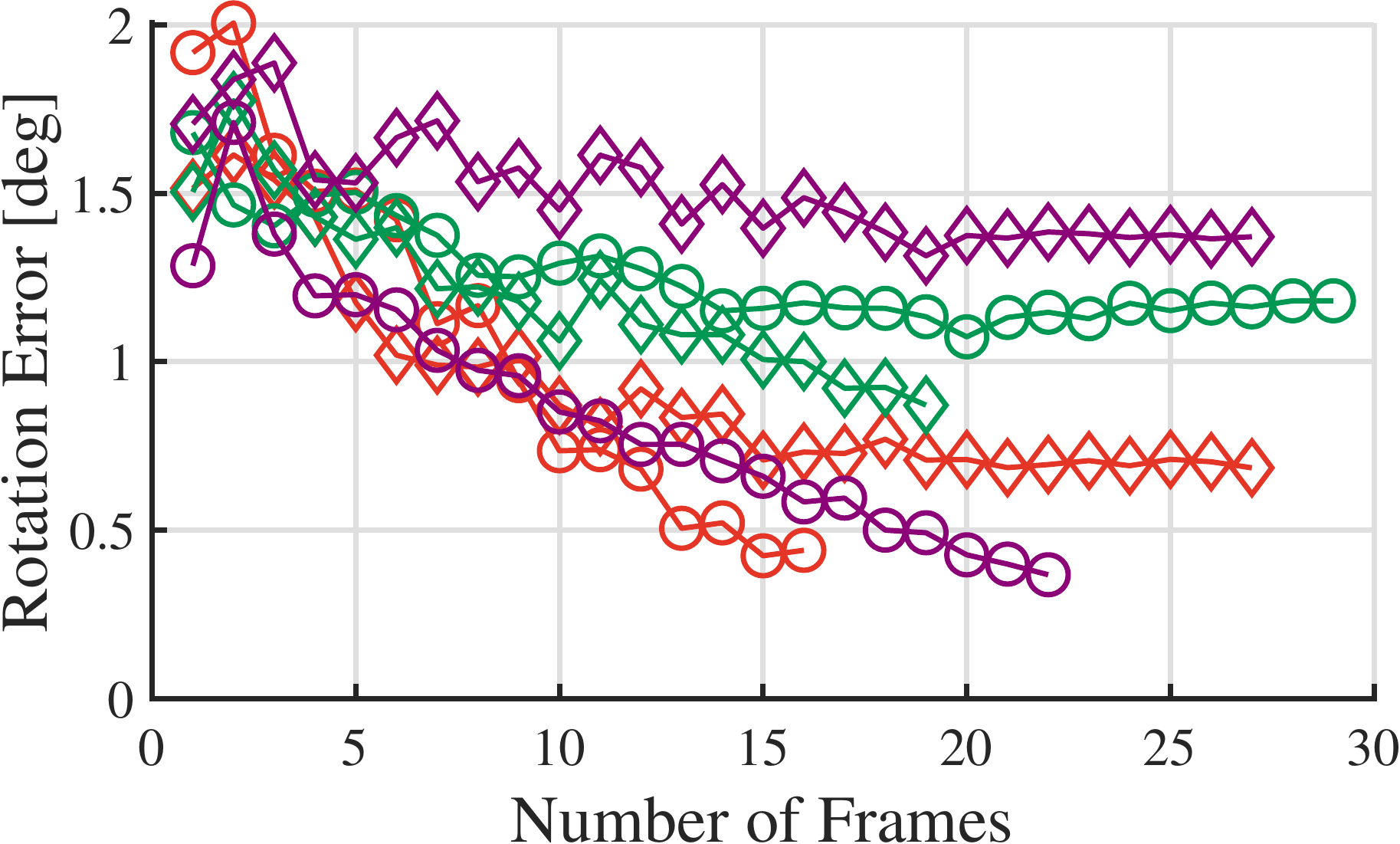}}
  \subfigure[Median translation error]
  {\label{fig:exp_rles_error_tsl}\centering\includegraphics[width=0.31\textwidth]{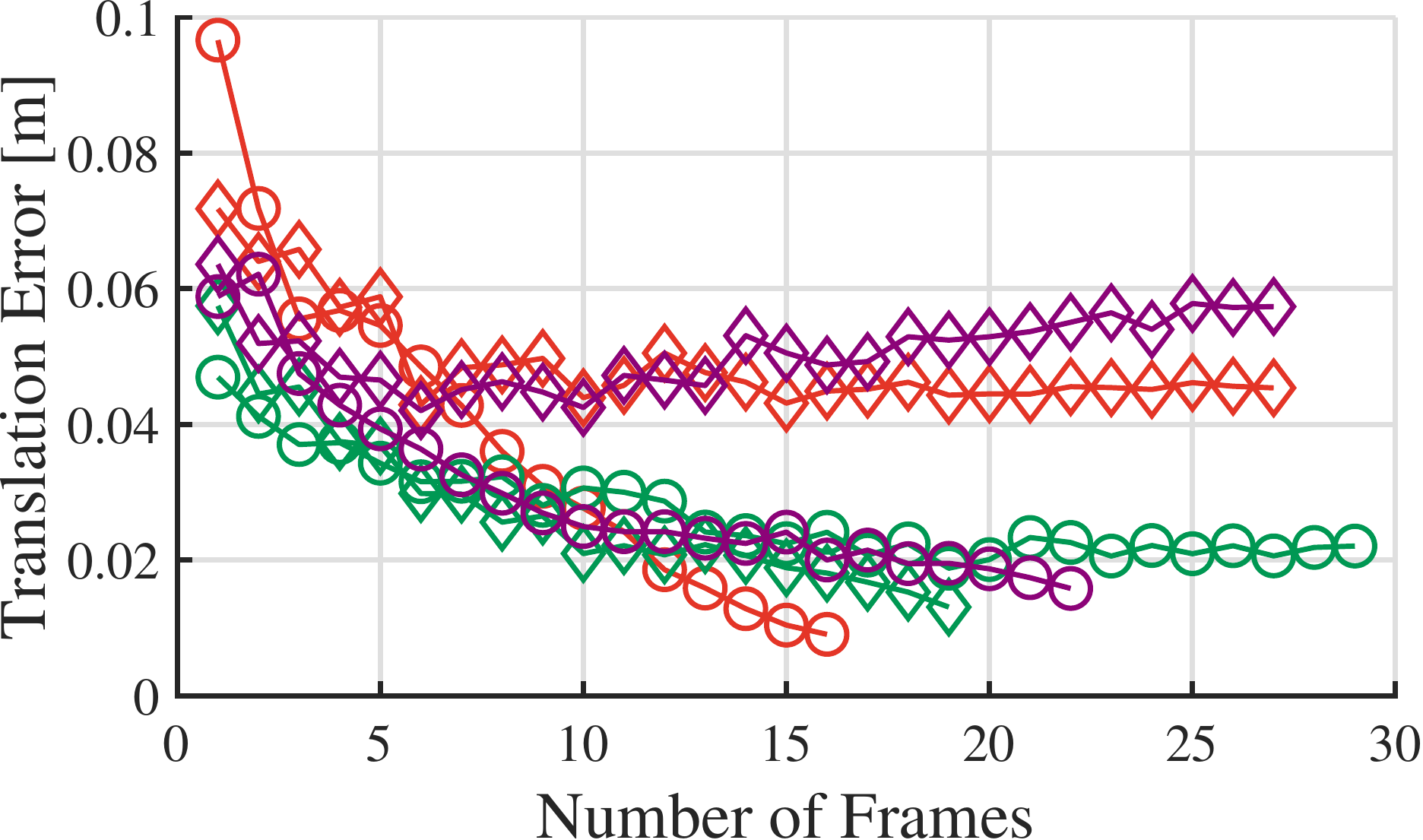}}
  \subfigure[Median MGE]
  {\label{fig:exp_rles_error_tsl}\centering\includegraphics[width=0.342\textwidth]{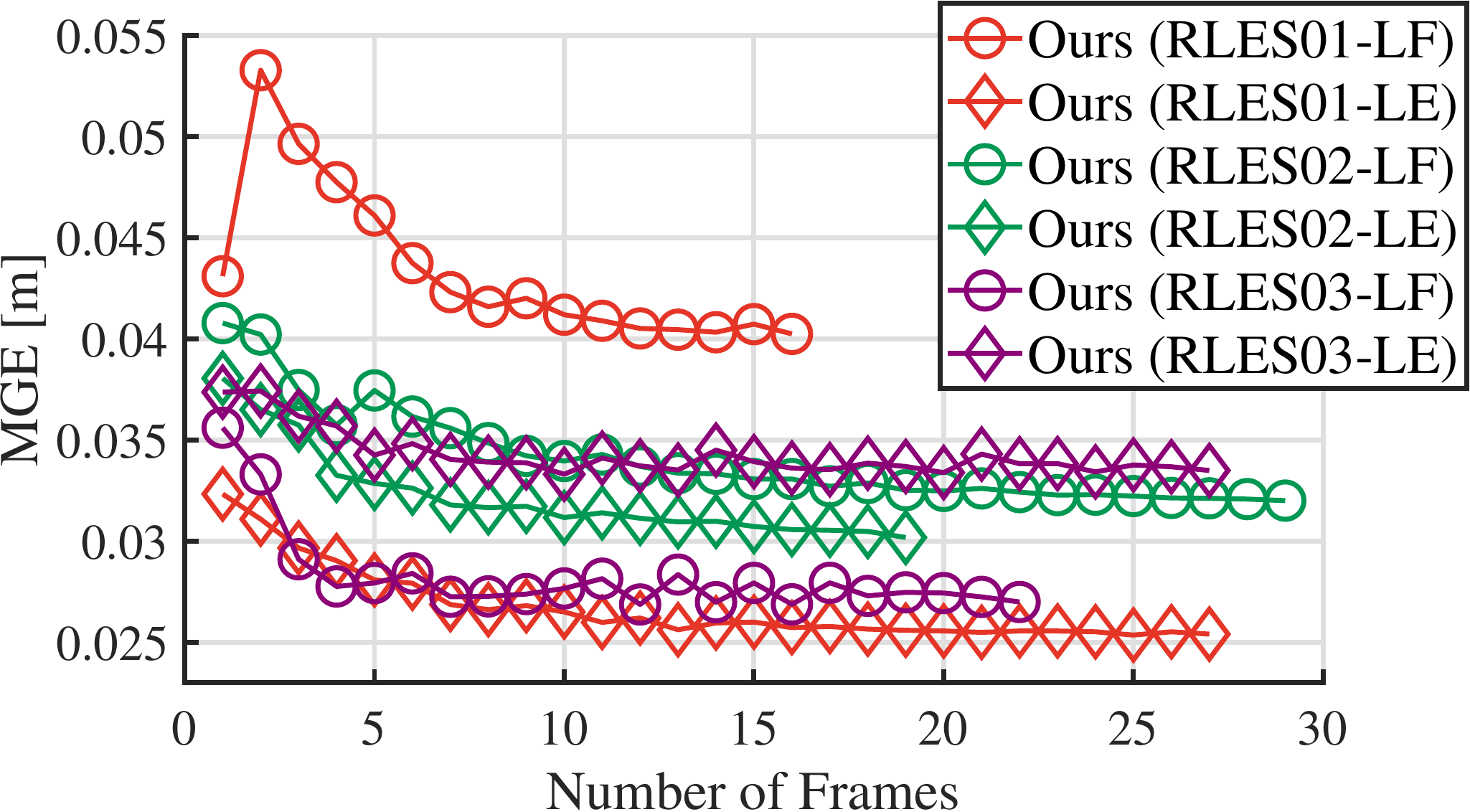}}
  \caption{Median rotation error, translation error, and MGE at multiple runs	of our method on calibration data: \textit{RLES01-LF}, \textit{RLES01-LE}, \textit{RLES02-LF}, \textit{RLES02-LE}, \textit{RLES03-LF}, and \textit{RLES03-LE}. The median errors are computed using the set of candidate extrinsics at line $8$ of Algorithm \ref{alg:ext_refinement}.}
  \label{fig:exp_rles_error}
\end{figure*}

\begin{table*}[]
  \centering
  \caption{Calibration results on \textit{RLES}. $\downarrow$/$\uparrow$ indicates that the lower/higher the value, the better the score.}
  \renewcommand\arraystretch{0.9}
  \renewcommand\tabcolsep{4.6pt}
  \scriptsize
  \begin{tabular}{ccccccccccccc}
    \toprule
    \multirow{2}{*}{Dataset}                               &
    \multirow{2}{*}{Method}                                &
    \multicolumn{4}{c}{Quaternion}                         &
    \multicolumn{3}{c}{Translation {$[m]$}}                &
    \multirow{2}{*}{$EGT_{\mathbf{R}}\ [deg, \downarrow]$} &
    \multirow{2}{*}{$EGT_{\mathbf{t}}\ [m, \downarrow]$}   &
    \multirow{2}{*}{\textit{MGE}$\ [m, \downarrow]$}       &
    \multirow{2}{*}{$N_{\text{detect}}/N\ [\uparrow]$}                                                                                                                                                                                                                                       \\
    \cline{3-9}
                                                           &                         & \multicolumn{1}{c}{$w$} & \multicolumn{1}{c}{$x$} & \multicolumn{1}{c}{$y$} & \multicolumn{1}{c}{$z$}
                                                           & \multicolumn{1}{c}{$x$} & \multicolumn{1}{c}{$y$} & \multicolumn{1}{c}{$z$} &                         &                         &                                                                                               \\
    \midrule[0.03cm]

                                                           & \textit{CAD}            & $\dgt{0.500}$           & $\dgt{0.500}$           & $\dgt{-0.500}$          & $\dgt{0.500}$           & $\dgt{0.126}$  & $\dgt{-0.065}$ & $\dgt{-0.080}$
                                                           & $-$                     & $-$                     & $-$                     & $-$                                                                                                                                               \\
    \midrule
    \midrule

                                                           & \textit{Fake GT}        &
    $\dgt{0.502}$                                          & $\dgt{0.504}$           & $\dgt{-0.498}$          & $\dgt{0.495}$           & $\dgt{0.112}$           & $\dgt{-0.094}$          & $\dgt{-0.108}$
                                                           & $-$                     & $-$                     & $-$                     & $-$                                                                                                                                               \\
    \midrule

    \multirow{2}{*}{\textit{RLES01-LF}}
                                                           & \textit{Ours}
                                                           & $0.499$                 & $0.505$                 & $-0.498$                & $0.498$                 & $0.126$                 & $-0.095$       & $-0.106$       & $\bm{0.629}$   & $\bm{0.032}$ & $0.037$      & $\bm{20}/20$
    \\
                                                           & \textit{Zhou-MATLAB}
                                                           & $0.502$                 & $0.498$                 & $-0.502$                & $0.498$                 & $0.132$                 & $-0.094$       & $-0.104$       & $1.151$        & $0.038$      & $\bm{0.036}$ & $\bm{20}/20$
    \\
    \midrule

    \multirow{2}{*}{\textit{RLES01-LE}}
                                                           & \textit{Ours}
                                                           & $0.497$                 & $0.501$                 & $-0.501$                & $0.500$                 & $0.150$                 & $-0.092$       & $-0.096$       & $\bm{0.839}$   & $\bm{0.041}$ & $0.024$      & $\bm{33}/36$
    \\
                                                           & \textit{Zhou-MATLAB}
                                                           & $0.496$                 & $0.500$                 & $-0.500$                & $0.503$                 & $0.161$                 & $-0.097$       & $-0.100$       & $1.258$        & $0.050$      & $\bm{0.023}$ & $\bm{33}/36$
    \\
    \midrule
    \midrule

                                                           & \textit{Fake GT}        &
    $\dgt{0.500}$                                          & $\dgt{0.505}$           & $\dgt{-0.498}$          & $\dgt{0.497}$           & $\dgt{0.129}$           & $\dgt{-0.080}$          & $\dgt{-0.109}$
                                                           & $-$                     & $-$                     & $-$                                                                                                                                                                         \\
    \midrule

    \multirow{2}{*}{\textit{RLES02-LF}}
                                                           & \textit{Ours}
                                                           & $0.497$                 & $0.506$                 & $-0.502$                & $0.495$                 & $0.131$                 & $-0.064$       & $-0.102$       & $\bm{0.677}$   & $\bm{0.018}$ & $\bm{0.031}$ & $\bm{34}/36$
    \\
                                                           & \textit{Zhou-MATLAB}
                                                           & $0.495$                 & $0.504$                 & $-0.504$                & $0.497$                 & $0.160$                 & $-0.060$       & $-0.090$       & $0.927$        & $0.041$      & $0.037$      & $32/36$
    \\
    \midrule

    \multirow{2}{*}{\textit{RLES02-LE}}
                                                           & \textit{Ours}
                                                           & $0.498$                 & $0.507$                 & $-0.499$                & $0.496$                 & $0.130$                 & $-0.071$       & $-0.107$       & $\bm{0.331}$   & $\bm{0.009}$ & $\bm{0.028}$ & $\bm{34}/36$
    \\
                                                           & \textit{Zhou-MATLAB}
                                                           & $0.491$                 & $0.505$                 & $-0.505$                & $0.498$                 & $0.152$                 & $-0.049$       & $-0.096$       & $1.270$        & $0.041$      & $0.034$      & $31/36$
    \\
    \midrule

    \multirow{2}{*}{\textit{RLES03-LF}}
                                                           & \textit{Ours}
                                                           & $0.497$                 & $0.511$                 & $-0.496$                & $0.496$                 & $0.116$                 & $-0.055$       & $-0.110$       & $\bm{0.781}$   & $\bm{0.028}$ & $0.027$      & $\bm{29}/36$
    \\
                                                           & \textit{Zhou-MATLAB}
                                                           & $0.508$                 & $0.504$                 & $-0.494$                & $0.495$                 & $0.100$                 & $-0.104$       & $-0.114$       & $1.034$        & $0.039$      & $\bm{0.024}$ & $26/36$
    \\
    \midrule

    \multirow{2}{*}{\textit{RLES03-LE}}
                                                           & \textit{Ours}
                                                           & $0.504$                 & $0.503$                 & $-0.498$                & $0.495$                 & $0.119$                 & $-0.090$       & $-0.108$       & $0.554$        & $\bm{0.014}$ & $0.030$      & $\bm{29}/36$
    \\
                                                           & \textit{Zhou-MATLAB}
                                                           & $0.497$                 & $0.507$                 & $-0.499$                & $0.497$                 & $0.124$                 & $-0.066$       & $-0.101$       & $\bm{0.401}$   & $0.016$      & $\bm{0.028}$ & $27/36$
    \\
    \bottomrule[0.03cm]
  \end{tabular}
  \label{tab:rles_calibration_results}
\end{table*}

\subsection{Calibration Results on Simulated Data}
\label{sec:exp_calib_simulate}
We first verify the LiDAR-frame camera calibration with synthetic data.
The sampled image and point cloud are shown in Fig. \ref{fig:exp_simu_sensor}.
The LiDAR and camera are given sufficient rotational offset: $[-70,-70,150]deg$ at roll, pitch and yaw respectively, as well as translational offset: $[0.4185, -0.3050, -0.1476]m$ at $x-$, $y-$, and $z-$ axes respectively.

We place the checkerboard in front of the sensor with sufficient rotation and translation and obtain
$\bm{30}$ pairs of noise-free LiDAR-camera data.
To test the robustness and limitations of our method, we further generate another $\bm{9}$ sets of data.
First, we add zero-mean Gaussian noise on depth measurements with an increasing standard deviation of
$\sigma\in\{0.8, 1.6, 2.4, 3.2, 4.0, 4.8, 6.0, 8.0, 10.0\}cm$ to the noise-free LiDAR data.
Second, we additionally process points which are next to the checkerboard's boundaries by adding another Gaussian noise.
This is because we want to mimic real-world data since the practical emmitted laser has a divergence angle rather than a strict line \cite{yuan2021pixel}.

In Fig. \ref{fig:exp_err_plot_simu_data}, we plot the calibration error.
Benefiting from the proposed feature extraction and globally optimization, our method overall has more accurate results than \textit{Zhou-MATLAB} even under large noise.
A key step in feature extraction is estimating the normal vector of the checkerboard plane from LiDARs.
Both noisy/outlier points and improper plane fitting may negatively affect the estimates.
This figure also shows the mean angular error between and ground-truth normals and estimated normals.
\textit{Zhou-MATLAB} cannot estimate accurate planar coefficients.
Regarding the optimization part, our method has rich planar and edge constraints to enforce the accuracy.
Also, the globally optimal solution in our method is essential to these results.
Moreover, we find that the unsupervised metric: \textit{MGE} is also suitable to evaluate calibration results, since its curve is consistent with the curves of \textit{EGT}.

\begin{figure}[]
  \centering
  \includegraphics[width=0.49\textwidth]{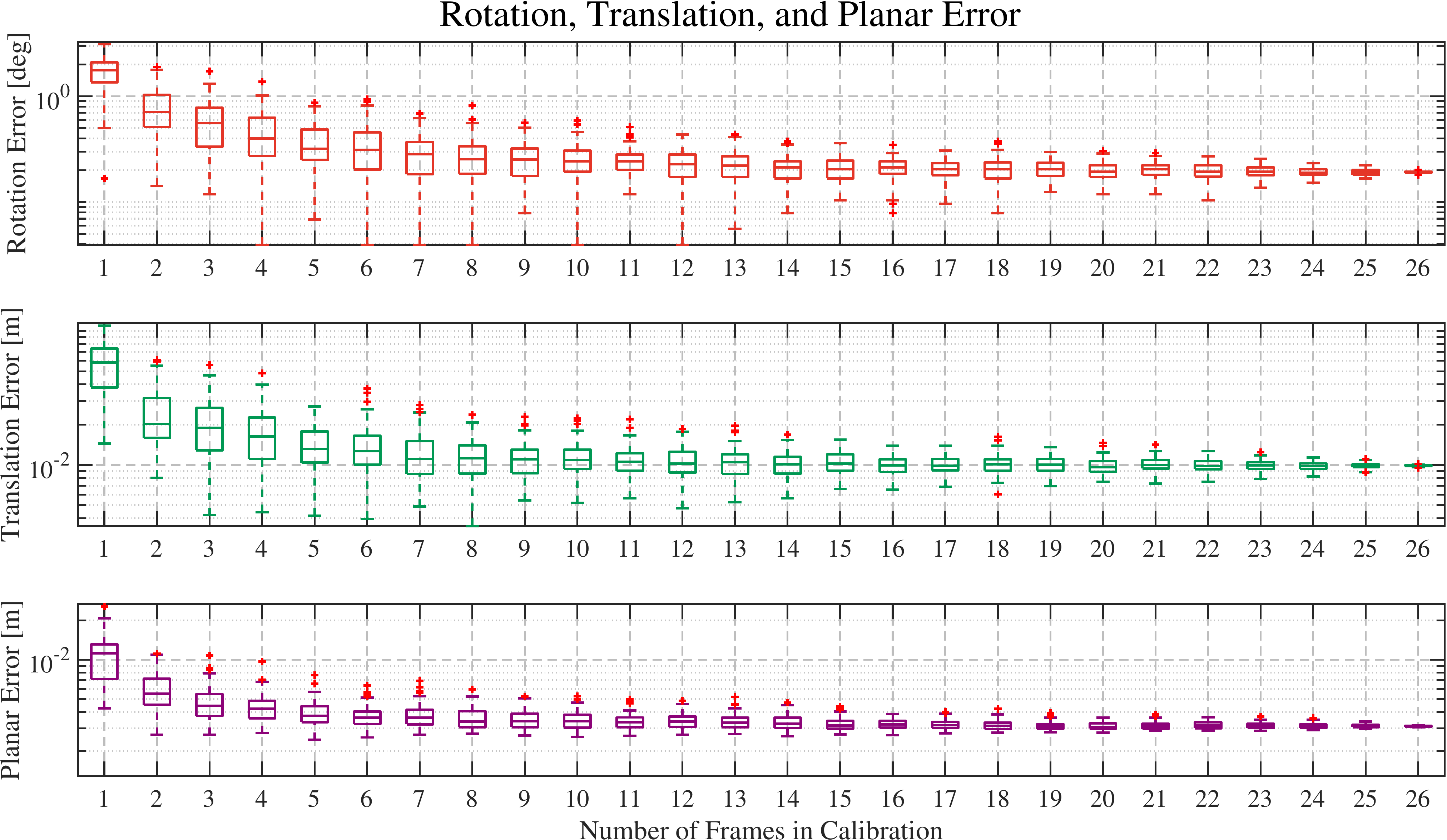}
  \caption{Calibration error along with frame number on \textit{RLFS01}.}
  \label{fig:calibration_error_frame_rlfs}
\end{figure}

\begin{figure}[t]
  \centering
  \subfigure[\textit{RLES01-LF}]
  {\label{fig:exp_rles_calibration_image_01_lf}\centering\includegraphics[width=0.22\textwidth]{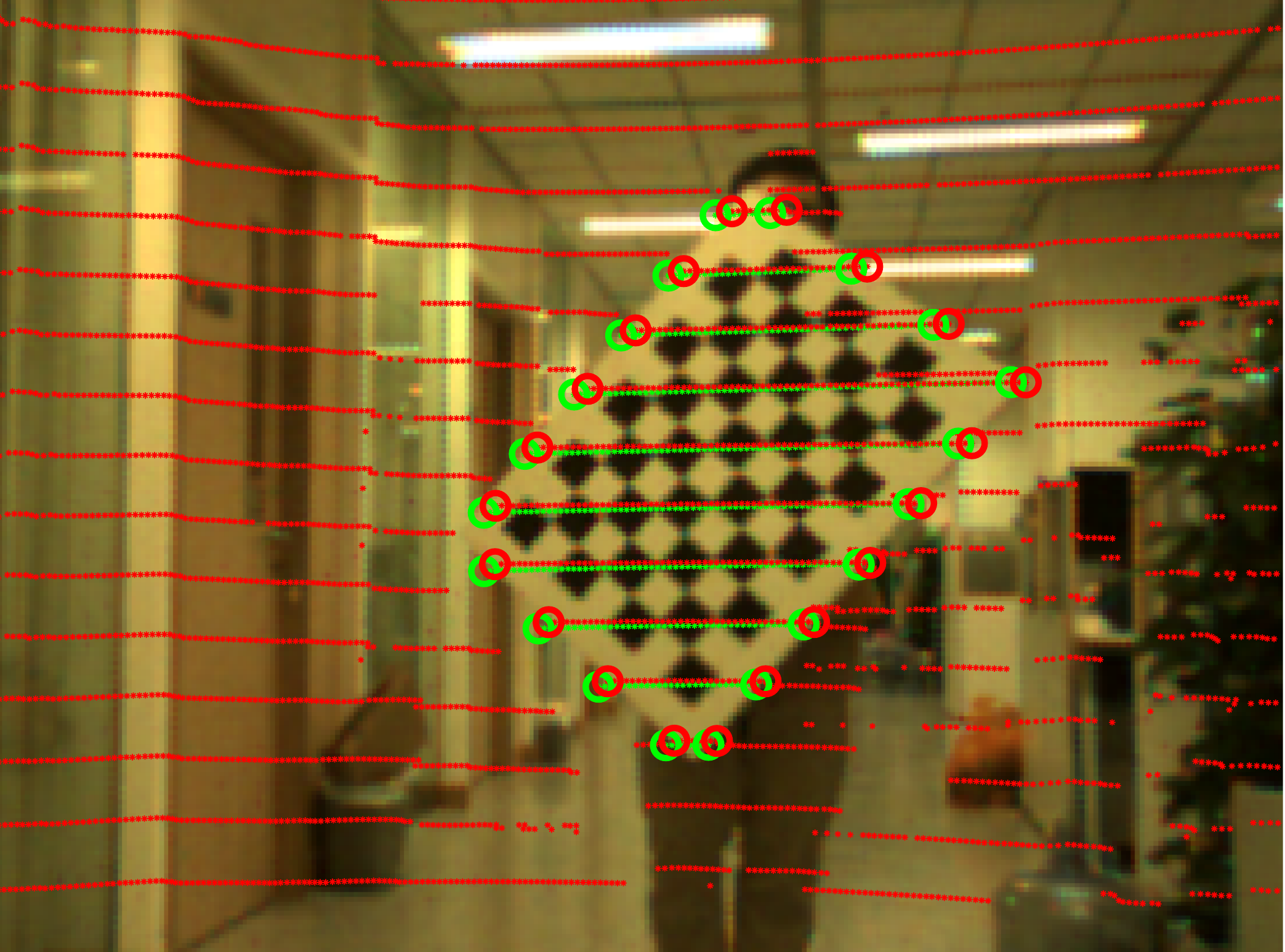}}
  \subfigure[\textit{RLES01-LE}]
  {\label{fig:exp_rles_calibration_image_01_le}\centering\includegraphics[width=0.22\textwidth]{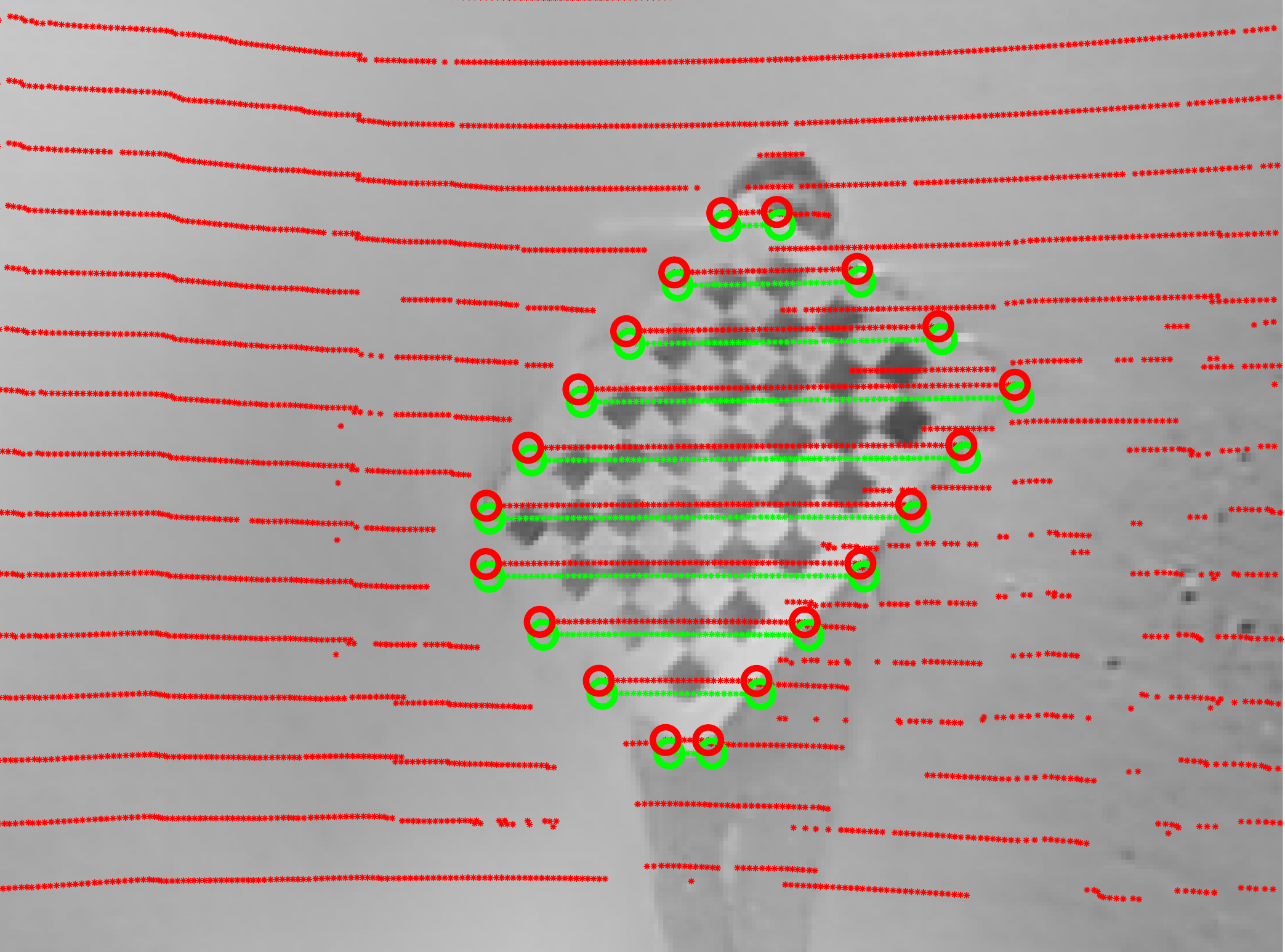}}
  \subfigure[\textit{RLES02-LF}]
  {\label{fig:exp_rles_calibration_image_02_lf}\centering\includegraphics[width=0.22\textwidth]{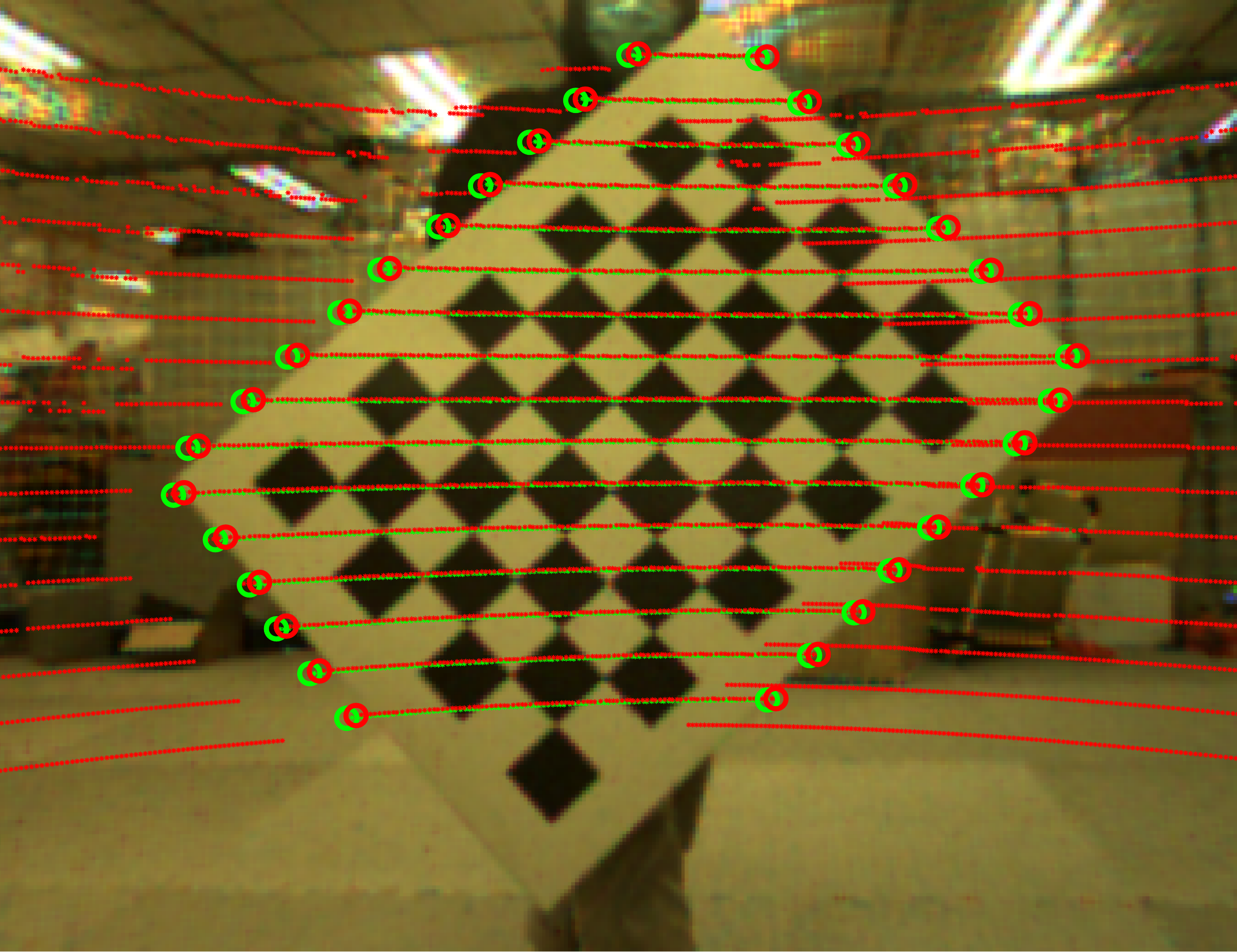}}
  \subfigure[\textit{RLES02-LE}]
  {\label{fig:exp_rles_calibration_image_02_le}\centering\includegraphics[width=0.22\textwidth]{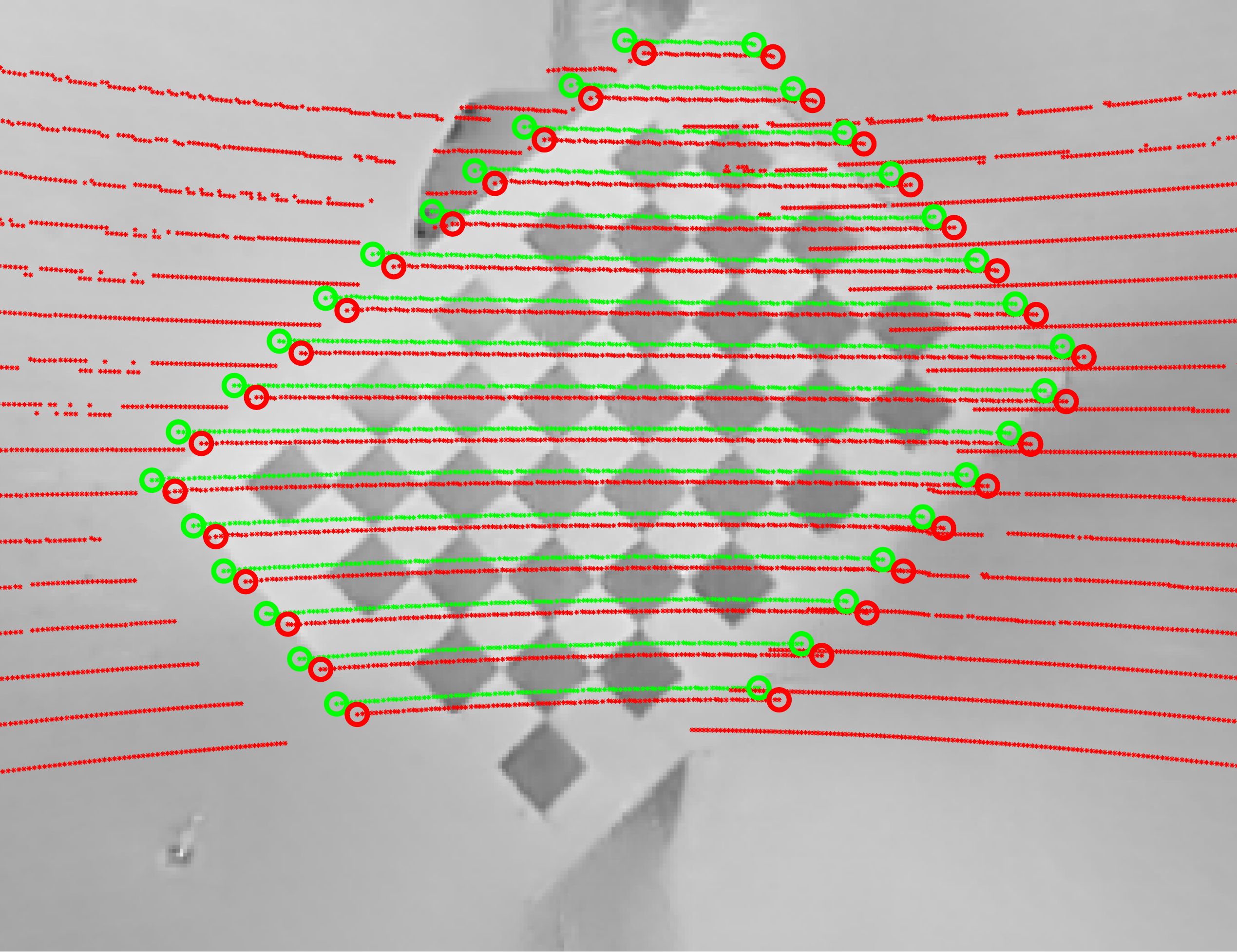}}
  \subfigure[\textit{RLES03-LF}]
  {\label{fig:exp_rles_calibration_image_03_lf}\centering\includegraphics[width=0.22\textwidth]{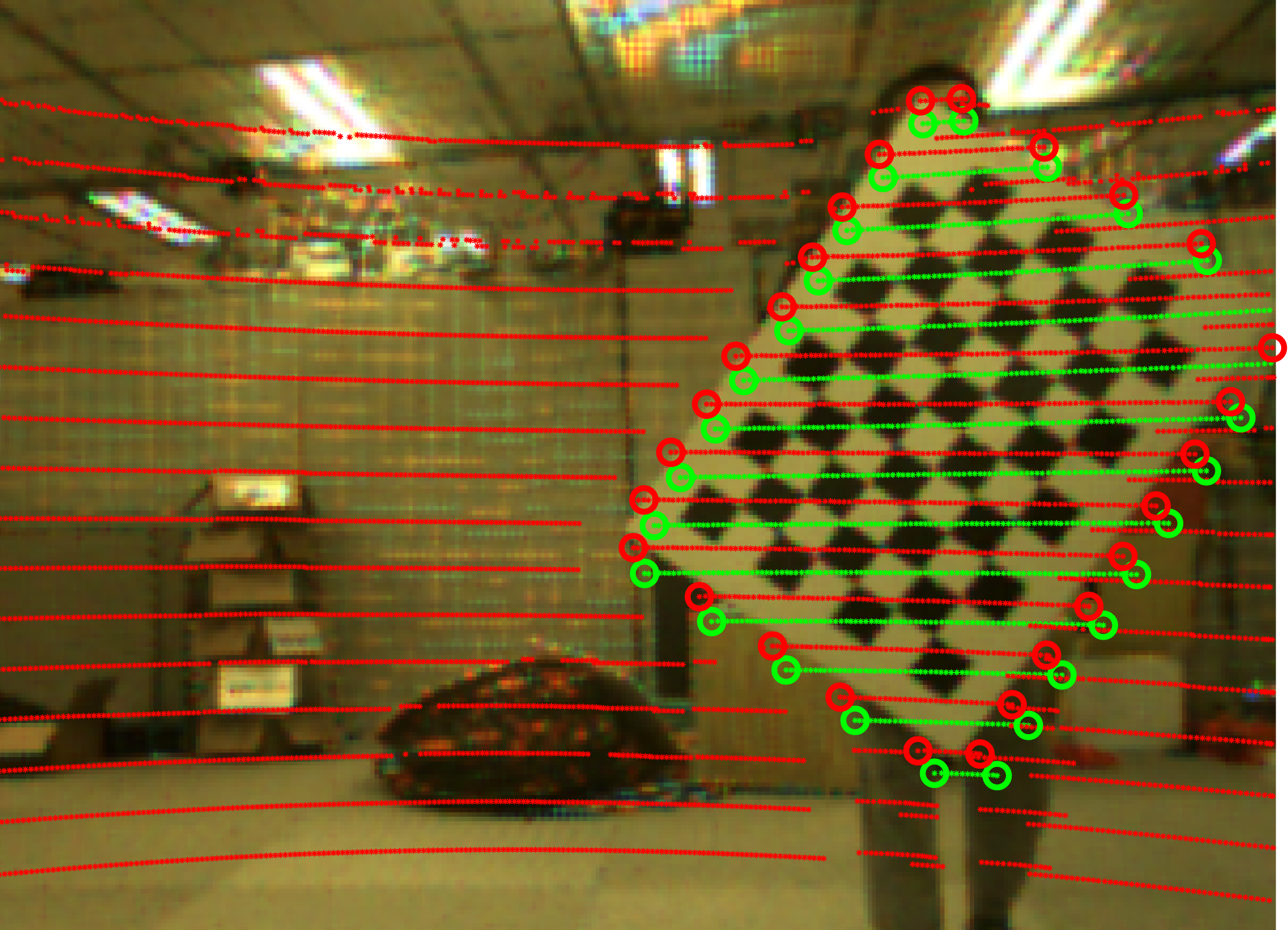}}
  \subfigure[\textit{RLES03-LE}]
  {\label{fig:exp_rles_calibration_image_03_le}\centering\includegraphics[width=0.22\textwidth]{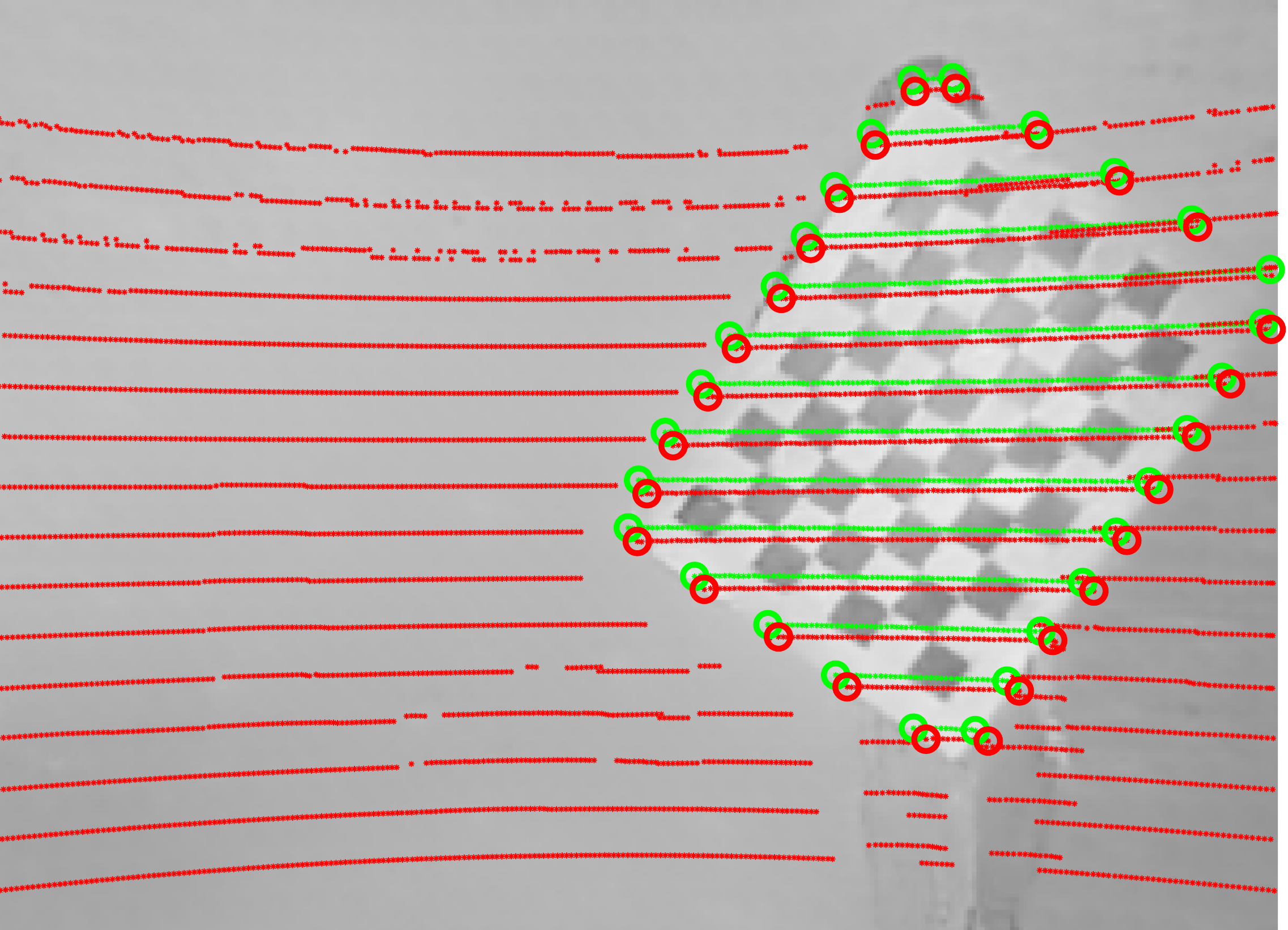}}
  \subfigure[\textit{RLES02} (cloud)]
  {\label{fig:exp_rles_calibration_02_cloud}\centering\includegraphics[width=0.22\textwidth]{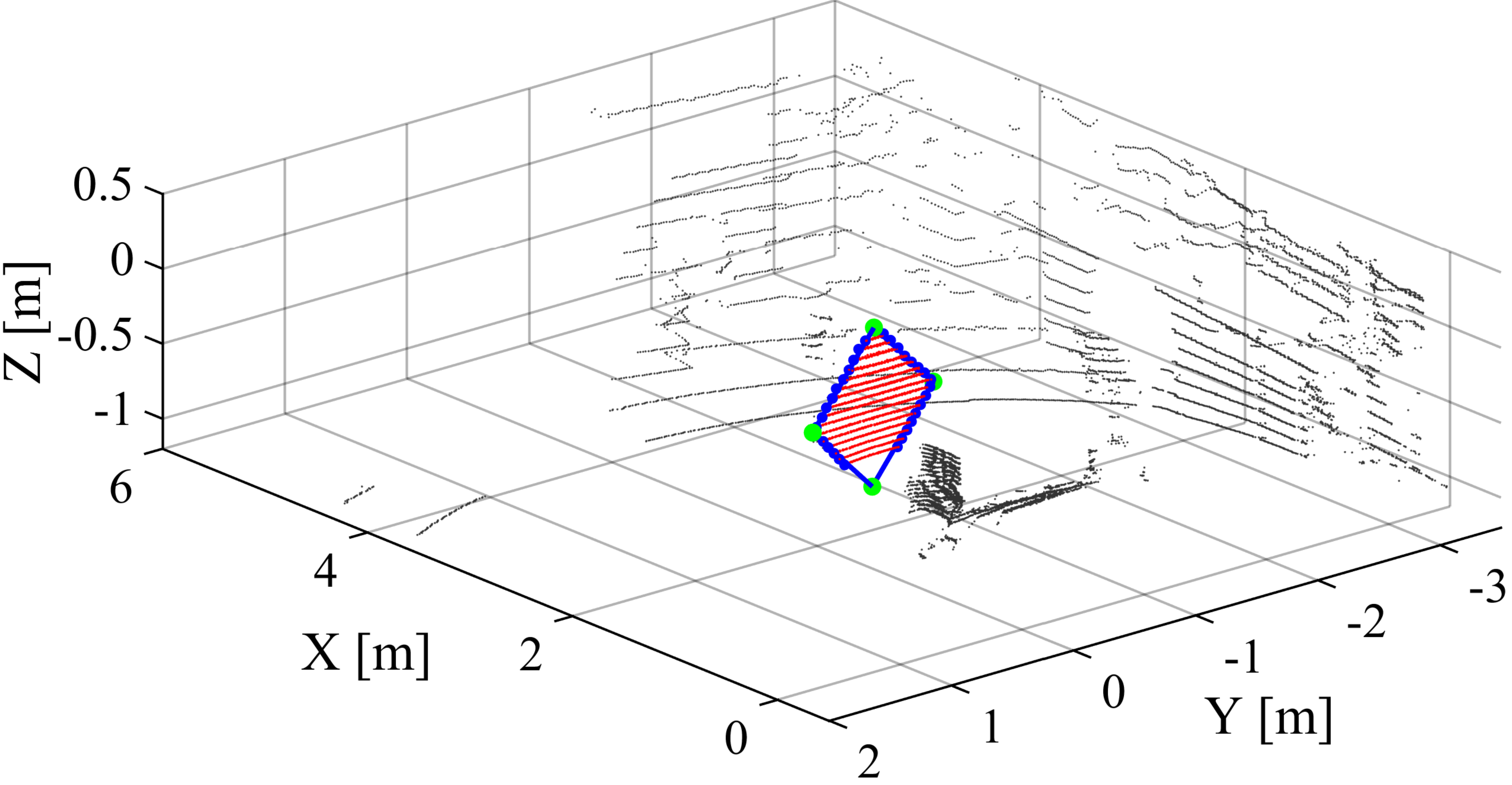}}
  \subfigure[\textit{RLES03} (cloud)]
  {\label{fig:exp_rles_calibration_03_cloud}\centering\includegraphics[width=0.22\textwidth]{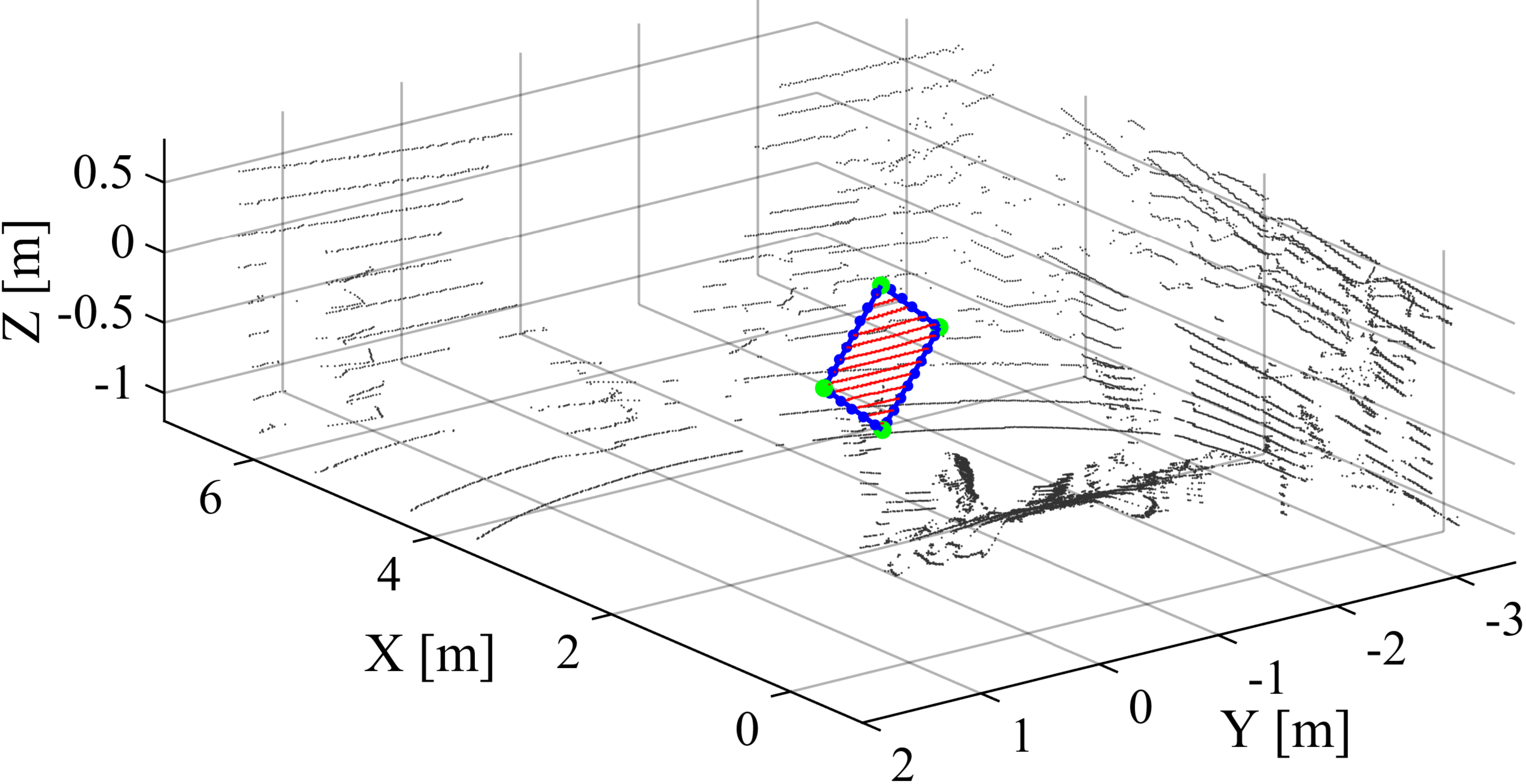}}
  \caption{Back-projection of the LiDAR points on images using the extrinsics in Table \ref{tab:rles_calibration_results}: our algorithm (red) and \textit{Zhou-MATLAB} (green).}
  \label{fig:exp_rles_calibration_image}
\end{figure}

\subsection{Calibration Results on Real-World Data}
\label{sec:exp_calib_realworld}

This section tests the proposed method with two real-world sensor suites.
Similarly, we move a checkerboard before sensors slowly to collect several groups of calibration data. Experimental results are shown in the following sections.

\subsubsection{Extrinsic Calibration of RLFS}
\label{sec:exp_rlfs}
We first verify the calibration method with the LiDAR-frame camera setup.
We collect three groups of data called \textit{RLFS01}, \textit{RLFS02}, and \textit{RLFS03} to test our approach.
\textit{RLFS01} and \textit{RLFS02} were collected in an indoor corridor. Besides the checkerboard, other objects such as grounds, walls, and square pillars appear as planar surfaces.
\textit{RLF03} were collected in an outdoor garden. Several noisy objects such as buildings, poles, and lawns exist, affecting the checkerboard detection from point clouds.

We firstly employ the MATLAB camera calibration toolbox\footnote{\url{https://www.mathworks.com/help/vision/camera-calibration.html}} to calibrate the camera's intrinsics with these collected data and then calibrate extrinsics.
Our method successfully recovers the extrinsics.
Fig. \ref{fig:exp_rlfs_error} plots errors with increasing frames of measurements and Table \ref{tab:rlfs_calibration_results} gives quantitative sample results.
$N_{\text{detect}}/N$ shows the portion of successfully detecting the checkerboard from images or point clouds.
\textit{Zhou-MATLAB} only uses a portion of measurements since it fails to detect checkerboards from several LiDAR frames and correctly detect features, thus obtaining inaccurate extrinsics.
These results are consistent with those in Section \ref{sec:exp_calib_simulate}, where our method outperforms the baseline and reaches up to $<0.3deg$ and $<1.5cm$ calibration error as well as $<0.9cm$ \textit{MGE} in both indoor and outdoor environments.

Fig. \ref{fig:exp_rlfs_calibration_image} shows the back-projection results using extrinsics in Table \ref{tab:rlfs_calibration_results}.
Circular points indicate detected edge points.
Our method can provide better results since the projections of edge points nearly stay on the the checkerboard's boundary.
We additionally show the calibration error along with the number of frames used in
calibration. The error gradually decreases since more constraints help to enforce the accuracy. One example on \textit{RLFS01} is shown in Fig. \ref{fig:calibration_error_frame_rlfs}.

\subsubsection{Extrinsic Calibration of RLES}
\label{sec:exp_rles}

We further verify the complete LCE-Calib method with RLES.
Three groups of calibration data are collected in an indoor room under different lighting conditions, which are called \textit{RLES01}, \textit{RLES02}, and \textit{RLES03} respectively.
In contrast to the last section, we additionally compare calibration results of using frame images (\textit{X-LF}) and reconstructed images (\textit{X-LE}) from event streams, respectively.
Fig. \ref{fig:exp_rles_error} plots calibration errors of the proposed method and Table \ref{tab:rles_calibration_results} reports quantitative results.

The calibration errors consistently decrease with more frames.
Though, compared with Table \ref{tab:rlfs_calibration_results}, the calibration errors are larger.
We consider that this is mainly caused by the low resolution of the event camera, affecting the inner pattern detection accuracy.
Cameras' low resolution also makes \textit{MGE} values not consistent with \textit{EGT} in some cases since boards' positions are not detected very precisely.
But we can still use this metric for selecting good extrinsics.
From Fig. \ref{fig:exp_rles_error}, we observe that extrinsics converge if around $\bm{15}$ frames are given. More data can further enforce accuracy.
For event cameras, inner patterns and boundaries of the board are reconstructed clearly. Thus, the calibration accuracy does not have much degradation.
A sufficient number of frames are used in calibration, whether using frame images or images reconstructed from event streams, and the extrinsics are successfully recovered.
Fig. \ref{fig:exp_rles_calibration_image} shows the back-projection results of LiDAR points using extrinsics in Table \ref{tab:rles_calibration_results}.
The checkerboard detection results from LiDAR points are also shown.
In \textit{RLES03-LF}, extrinsics from our method can help LiDAR points better align on the board's boundaries than \textit{Zhou-MATLAB}.

\begin{table}[t]
	\centering
	\caption{Calibration results on \textit{RLFS} with our proposed method by disabling some modules. $\downarrow$/$\uparrow$ indicates that the lower/higher the value, the better the score. The first two results are marked as bold.}
	\renewcommand\arraystretch{1.0}
	\renewcommand\tabcolsep{6.0pt}
	\scriptsize
	\begin{tabular}{ccccc}
		\toprule
		\multirow{1}{*}{Dataset}                              &
		\multirow{1}{*}{Method}                               &
		\multirow{1}{*}{$EGT_{\mathbf{R}}\ [deg,\downarrow]$} &
		\multirow{1}{*}{$EGT_{\mathbf{t}}\ [m,\downarrow]$}   &
		\multirow{1}{*}{\textit{MGE}$\ [m,\downarrow]$}
		\\
		\midrule[0.03cm]

		\multirow{7}{*}{\textit{RLFS01}}

		                                                      & \textit{WO. PP}   & $0.517$      & $0.023$      & $0.015$      \\
		                                                      & \textit{WO. UM}   & $\bm{0.119}$ & $0.010$      & $\bm{0.008}$ \\
		                                                      & \textit{WO. PtPL} & $0.292$      & $0.011$      & $\bm{0.008}$ \\
		                                                      & \textit{WO. PtL}  & $0.556$      & $0.018$      & $\bm{0.008}$ \\
		                                                      & $I=1$             & $0.195$      & $\bm{0.007}$ & $0.009$      \\
		                                                      & $I=5$             & $0.262$      & $0.012$      & $\bm{0.008}$ \\
		                                                      & \textit{Ours}     & $\bm{0.130}$ & $\bm{0.009}$ & $0.009$      \\
		\midrule

		\multirow{7}{*}{\textit{RLFS02}}

		                                                      & \textit{WO. PP}   & $0.369$      & $0.021$      & $0.015$      \\
		                                                      & \textit{WO. UM}   & $0.338$      & $0.015$      & $0.008$      \\
		                                                      & \textit{WO. PtPL} & $0.314$      & $\bm{0.012}$ & $0.008$      \\
		                                                      & \textit{WO. PtL}  & $0.653$      & $0.021$      & $\bm{0.007}$ \\
		                                                      & $I=1$             & $\bm{0.277}$ & $\bm{0.014}$ & $0.008$      \\
		                                                      & $I=5$             & $0.342$      & $0.015$      & $0.008$      \\
		                                                      & \textit{Ours}     & $\bm{0.293}$ & $\bm{0.014}$ & $\bm{0.007}$ \\
		\midrule

		\multirow{7}{*}{\textit{RLFS03}}

		                                                      & \textit{WO. PP}   & $0.378$      & $0.018$      & $0.016$      \\
		                                                      & \textit{WO. UM}   & $\bm{0.207}$ & $\bm{0.007}$ & $\bm{0.008}$ \\
		                                                      & \textit{WO. PtPL} & $0.338$      & $0.016$      & $0.009$      \\
		                                                      & \textit{WO. PtL}  & $0.289$      & $0.019$      & $\bm{0.007}$ \\
		                                                      & $I=1$             & $0.475$      & $0.024$      & $\bm{0.008}$ \\
		                                                      & $I=5$             & $0.324$      & $0.017$      & $\bm{0.008}$ \\
		                                                      & \textit{Ours}     & $\bm{0.136}$ & $\bm{0.008}$ & $\bm{0.008}$ \\
		\bottomrule[0.03cm]
		\multicolumn{5}{l}{WO.: without. PP: point projection. UM: uncertainty modelling of normal vectors.}                   \\
		\multicolumn{5}{l}{PtPL: point-to-plane constriants. PtL: point-to-line constriants.}                                  \\
		\multicolumn{5}{l}{$I$: the computation number in Algorithm \ref{alg:ext_refinement}.}                                 \\
	\end{tabular}
	\label{tab:ablation_study}
\end{table}

\subsection{Ablation Study}
\label{sec:exp_ablation_study}

Our method proposes several steps to handle the uncertainty of sensor data and calibration results, such as the point projection (Section \ref{sec:methodology_fsl_feature_extraction}), uncertainty modelling of normal vectors (Section \ref{sec:methodology_fsc}), and calibration with multiple computations (Section \ref{sec:ext_calibration}).
To show the impact of these individual modules, we conduct the following ablation study.
We disable one of these modules and repeatedly test the proposed method with \textit{RLFS}.
Table \ref{tab:ablation_study} reports the calibration accuracy.

Disabling the point projection \textit{(WO. PP)} or PtL constraints \textit{(WO. PtL)} significantly degrades the calibration performance.
This is because the point projection reduces noise of LiDAR points with the fitted planar parameters and the checkerboard's edges (the source of PtL constraints) offer strong constraints to extrinsics.
We also observe that setting $I=1$ or $I=5$ may sometimes induce unreliable results (see \textit{RLFS03}).
To minimize the error bound of the results, the value of $I$ cannot be small in practice.

\section{Conclusion}
\label{sec:conclusion}


This paper presents a novel automatic checkerboard-based approach to calibrate extrinsics between a LiDAR and a frame/event camera, which offers several desirable features such as automatic checkerboard detection and tracking, image reconstruction from event streams, and globally optimal extrinsics estimation. The proposed method has been extensively evaluated on both simulated sensors and real-world devices, demonstrating its superior performance in terms of accuracy in translation and rotation compared to a SOTA checkerboard-based method.
Moving forward, future research could explore the effect of time undistortion on camera and LiDAR sensors, as well as further advancing the LCE configuration to address various robotics challenges, such as ego-motion estimation or object detection in complex scenarios.


\bibliographystyle{IEEEtran}
\bibliography{reference}{}

\end{document}